\documentclass[conference]{IEEEtran}
\IEEEoverridecommandlockouts
\usepackage{comment}
\usepackage{verbatim}
\usepackage{booktabs}
\usepackage{cite}
\usepackage{amsmath}
\usepackage{multirow}
\usepackage{amsmath,amssymb,amsfonts}
\usepackage{algorithmic}
\usepackage{graphicx}
\usepackage{textcomp}
\usepackage{array}
\usepackage[linesnumbered,ruled,vlined]{algorithm2e}
\usepackage{romannum}
\usepackage{xcolor}
\def\BibTeX{{\rm B\kern-.05em{\sc i\kern-.025em b}\kern-.08em
    T\kern-.1667em\lower.7ex\hbox{E}\kern-.125emX}}
\begin{document}
\title{Boosting Federated Domain Generalization: Understanding the Role of Advanced Pre-Trained Architectures \\

 \thanks{*Corresponding author: Choong Seon Hong (cshong@khu.ac.kr)}
}
\author{
\IEEEauthorblockN{Avi Deb Raha\IEEEauthorrefmark{1}, Apurba Adhikary\IEEEauthorrefmark{1}, Mrityunjoy Gain\IEEEauthorrefmark{2}, Yu Qiao\IEEEauthorrefmark{2}, and Choong Seon Hong\IEEEauthorrefmark{1*}}
\IEEEauthorblockA{\IEEEauthorrefmark{1}\textit{Department of Computer Science and Engineering, Kyung Hee University, Yongin-si 17104, Republic of Korea}\\
	E-mail: avi@khu.ac.kr, apurba@khu.ac.kr, cshong@khu.ac.kr}
\IEEEauthorblockA{\IEEEauthorrefmark{2}\textit{Department of Artificial Intelligence, Kyung Hee University, Yongin-si 17104, Republic of Korea}\\
	E-mail: gain@khu.ac.kr, qiaoyu@khu.ac.kr}
}
\maketitle
\begin{abstract}
Federated learning enables collaborative model training without data sharing, ensuring privacy. However, the heterogeneity of client data poses significant challenges to model generalization across diverse data distributions. Existing federated domain generalization (FDG) methods predominantly rely on ResNet backbones pre-trained on ImageNet-1K, which may limit model adaptability due to both their architectural constraints and the limitations of the pre-training dataset. In this study, we explore the efficacy of advanced pre-trained architectures, such as Vision Transformers (ViT), ConvNeXt, and Swin Transformers in enhancing FDG. These architectures capture global contextual features and model long-range dependencies, making them promising candidates for improving cross-domain generalization. We conduct a broad study with in-depth analysis and systematically evaluate different variants of these architectures, using extensive pre-training datasets such as ImageNet-1K, ImageNet-21K, JFT-300M, and ImageNet-22K. Additionally, we compare self-supervised and supervised pre-training strategies to assess their impact on FDG performance. Our findings suggest that self-supervised techniques, which focus on reconstructing masked image patches, can better capture the intrinsic structure of images, thereby outperforming their supervised counterparts. Comprehensive evaluations on the Office-Home and PACS datasets demonstrate that adopting advanced architectures pre-trained on larger datasets establishes new benchmarks, achieving average accuracies of 84.46\% and 92.55\%, respectively. Additionally, we observe that certain variants of these advanced models, despite having fewer parameters, outperform larger ResNet models. This highlights the critical role of utilizing sophisticated architectures and diverse pre-training strategies to enhance FDG performance, especially in scenarios with limited computational resources where model efficiency is crucial. Our results indicate that federated learning systems can become more adaptable and efficient by leveraging these advanced methods, offering valuable insights for future research in FDG.
\end{abstract}


\section{Introduction}

Deep learning has revolutionized numerous fields by enabling the extraction of complex patterns from vast amounts of data. However, its reliance on large centralized datasets raises significant concerns about data privacy, security, and the logistical challenges of data aggregation \cite{1_liu2020privacy}. Moreover, deep learning models often struggle with scenarios that differ from their training data, limiting their effectiveness in applications requiring adaptability to diverse and evolving environments \cite{raha2024advancing, 6_zhou2021domain, 8_kim2022broad}. Real-world data is also frequently distributed across multiple clients or devices rather than being centralized \cite{2_qiao2023mp, 3_liu2021feddg, 4_chen2023federated, 5_wu2021collaborative, park2024stablefdg, 9_mendieta2022local}, posing significant challenges for traditional approaches. This distribution complicates data collection, training processes, and model updates while exacerbating privacy and security concerns.

Federated Learning (FL) has emerged as a promising decentralized learning paradigm to address these challenges \cite{15_fedavg}. It enables the training of models across multiple devices or servers holding local data samples without sharing raw data \cite{2_qiao2023mp, 15_fedavg}. By keeping sensitive information on local devices, FL effectively mitigates the privacy and security issues inherent in traditional centralized methods. By aggregating model parameters derived from local data, FL facilitates a collaborative learning process that leverages the strengths of deep learning while addressing its limitations.

However, a common limitation in existing FL research is the assumption of consistent data distributions during training and testing \cite{2_qiao2023mp, 10_wu2023faster, 11_huang2023rethinking, 15_fedavg}. This assumption rarely reflects real-world conditions, where domain shifts are common and can significantly degrade model performance. For instance, FL clients may have data from specific source domains, such as images under sunny and rainy conditions, but the globally trained model must also perform accurately on data from unseen target domains like snowy conditions. This scenario underscores the need to integrate domain generalization (DG) capabilities into FL systems to ensure their effectiveness across diverse and evolving data environments \cite{IoT_01, IoT_02, 10_wu2023faster, 11_huang2023rethinking}.

DG is a sophisticated approach to developing models capable of maintaining high performance across new, unseen domains by utilizing data from known domains \cite{3_liu2021feddg, 4_chen2023federated, 5_wu2021collaborative, 6_zhou2021domain, park2024stablefdg, 8_kim2022broad, IoT_03}. This method has proven successful in centralised systems where it is possible to combine data from multiple domains for comprehensive model training. However, the transition of DG methods to the FL paradigm introduces unique challenges that complicate their direct application.
In the FL setup, the decentralized nature of data across clients presents significant challenges for achieving effective DG \cite{3_liu2021feddg, 5_wu2021collaborative, park2024stablefdg}. Each client typically has access to only a fragmented portion of the overall dataset, which may be limited not only in size but also in the diversity of data styles and characteristics. This limitation hinders local models from learning features that generalize well across different settings, which is crucial for DG. As a result, models trained on such limited and non-diverse data struggle to perform well in new, real-world situations. Addressing this limitation is essential for these models to adapt effectively to new domains.

Although several researchers have recently focused on DG for FL \cite{2_qiao2023mp, 3_liu2021feddg, 4_chen2023federated, 5_wu2021collaborative, 6_zhou2021domain}, most of these works utilize ResNet-50 or ResNet-18 \cite{12_he2016deep} pre-trained on ImageNet-1K \cite{36_russakovsky2015imagenet} as the backbone architecture. However, more advanced architectures such as Vision Transformers (ViT) \cite{dosovitskiy2020image}, Swin Transformers \cite{13_liu2021swin}, and ConvNeXt \cite{ConvNeXt} have emerged recently. These architectures have shown promising results on various classification and object detection tasks. Furthermore, they can learn robust features \cite{paul2022vision, bai2021transformers, zhang2024ci}, which may provide better generalization capabilities for the global model in Federated Domain Generalization (FDG). Moreover, a recent study \cite{8_kim2022broad} found that strong pre-training on broader datasets such as ImageNet-22K \cite{36_russakovsky2015imagenet} or JFT-300M \cite{jft-300m} can outperform existing state-of-the-art (SOTA) models in the centralized paradigm.

Despite advancements in neural network architectures and their success in centralized settings, there is a notable gap in exploring these advanced models (e.g., ViT, Swin Transformer, ConvNeXt) pre-trained on larger datasets for FDG. This oversight presents an opportunity to utilize stronger pre-trained models within the FDG context. These architectures are designed to be more adaptive, which is essential for handling the heterogeneous and distributed nature of data in FL scenarios. For instance, the self-attention mechanisms in Transformers enable dynamic recalibration of feature responses based on input data, potentially offering enhanced feature extraction across diverse domains unseen during initial training. By leveraging these advanced architectures, which have demonstrated substantial performance improvements, we can address the dual challenge of ensuring privacy and enhancing the robustness and generalizability of models in diverse, decentralized data environments. Therefore, this study explores the effects of next-generation models pre-trained on larger datasets within the FDG setup.

The following are the key contributions of this study:
\begin{itemize}

\item  We conduct the first comprehensive investigation of the effects of pre-trained architectures on FDG. Our research explores the impact of next-generation architectures within the challenging context of FDG. We assess their performance across a wide variety of pre-training datasets and compare self-supervised versus supervised pre-training practices. This study demonstrates how these advanced methods can enhance the robustness and effectiveness of models in FL environments.

\item  Several advanced architectures and their variants, including the Swin Transformer, ViT, and ConvNeXt, have been explored in the context of FDG. These models excel at capturing complex patterns, long-range dependencies, and hierarchical data representations, making them well-suited for  domain generalization tasks in federated settings. Our investigation provides fresh insights into the performance enhancements achievable with these advanced architectures, highlighting their potential over traditional models.

\item  We analyze the performance of these advanced models using a range of pre-training datasets, including ImageNet-1K, ImageNet-12K, ImageNet-21K, ImageNet-22K, and JFT-300M. This comparative study investigates the benefits of leveraging larger and more diverse datasets, an aspect underutilized in previous FDG research. Our findings reveal how the scale and diversity of pre-training data influence the generalization capabilities of models in federated settings.

\item  We investigate the impact of self-supervised versus supervised pre-training strategies on FDG performance. This analysis provides insights into how different pre-training approaches influence learning dynamics and generalization, particularly across datasets with varying characteristics. Our results offer guidance on selecting suitable pre-training methods to optimize model performance in federated environments. Notably, we find that SSL pre-training using a masked image patch reconstruction strategy outperforms their supervised counterparts in FDG.

\item Our evaluations on the Office-Home and PACS datasets demonstrate that using advanced architectures pre-trained on larger datasets sets new performance benchmarks, achieving average accuracies of 84.46\% and 92.55\%, respectively. Moreover, the results show that certain variants of the advanced models, despite having fewer parameters, outperform larger ResNet models. This underscores the importance of employing sophisticated architectures and diverse pre-training strategies to enhance FDG performance, particularly under resource constraints where efficiency and model capacity are critical.

\end{itemize}
The remainder of this paper is organized as follows: Section \ref{releted_works} describes the releted works, followed by methodology in Section \ref{methodology}. Subsequently, Section \ref{experiment} presents the experimental setup and simulation results. Finally, the concluding remarks are provided in Section \ref{conclusion}.

\section{Related Works} \label{releted_works}
In this section, we first review recent advancements in deep learning network architectures, including both  Convolutional Neural Networks (CNNs) and transformer-based models. Next, we discuss several widely used pre-training datasets and their influence on model performance and generalization capabilities. We then explore state-of-the-art (SOTA) self-supervised learning (SSL) techniques that enable models to learn from unlabeled data. Following that, we provide an overview of domain generalization (DG) methods. Finally, we conclude the related work section by introducing FL and FDG techniques, highlighting their significance in privacy-preserving distributed learning and their challenges in handling domain shifts.

\subsection{Network Architectures}\label{NA}
The effectiveness of deep learning models in performing various tasks is significantly influenced by their architectural design. CNNs have been the cornerstone of many SOTA applications since the introduction of AlexNet \cite{krizhevsky2012imagenet}. Subsequent advancements in CNN architectures have introduced more sophisticated features, such as deeper and more efficient convolutional layers. Notable examples include VGGNet \cite{simonyan2014very}, ResNet \cite{12_he2016deep}, SENet \cite{hu2018squeeze}, MobileNet \cite{howard2017mobilenets}, EfficientNet \cite{tan2019efficientnet}, and ConvNeXt \cite{ConvNeXt}.

In addition to CNNs, Transformer-based architectures have gained significant popularity due to their use of attention mechanisms, initially developed for natural language processing (NLP) \cite{vaswani2017attention}. Vision Transformers (ViT) \cite{dosovitskiy2020image} have demonstrated promising results by adapting these attention mechanisms for vision tasks, enhancing model capabilities. The introduction of Data-Efficient Image Transformers (DeiT) \cite{DEIT} further improved the training efficiency of ViTs. Similarly, the Swin Transformer \cite{13_liu2021swin} features a hierarchical design and employs sliding windows for self-attention, leading to superior performance in various computer vision tasks.
Further research \cite{bai2021transformers} indicates that Transformers can outperform traditional CNNs in terms of generalization on out-of-distribution samples, highlighting their robustness in real-world settings. Given their superior performance and generalization capabilities, advanced architectures such as ConvNeXt, ViT, and Swin Transformer are well-suited for Federated Domain Generalization (FDG). These architectures leverage attention mechanisms and hierarchical structures to handle diverse data sources effectively, maintaining consistent model performance across various contexts. Consequently, it is essential to examine the potential of these advanced architectures within the context of FDG.

\subsection{Pre-training Datasets}\label{PD}
The efficacy of deep learning models on downstream tasks is significantly influenced by the quality and scale of the datasets used during pre-training, particularly for vision-based applications \cite{kornblith2019better}. The ImageNet-1K \cite{36_russakovsky2015imagenet} dataset, comprising 1,000 classes and over one million images, has been instrumental in training numerous CNN models. This extensive dataset has enabled the development of robust pre-trained models that perform well across a variety of downstream tasks after fine-tuning.
Expanding the scope, the ImageNet-21K dataset includes approximately 21,841 categories and around 14 million images, offering greater diversity. The ImageNet-22K dataset, which includes over 22,000 classes, provides an even richer set of images for training, enhancing model generalization capabilities.

As the demand for models capable of handling a broader array of tasks increases, the utilization of more extensive and diverse datasets for pre-training is being increasingly explored. For example, Meta introduced the SA-1B dataset \cite{Kirillov_2023_ICCV}, which contains 11 million images and 1.1 billion mask annotations. This dataset serves as a specialized resource for training segmentation models, with masks presented in the COCO run-length encoding format without class labels. By focusing on precise object boundary delineation, SA-1B complements the extensive categorization of ImageNet, emphasizing spatial accuracy required in advanced vision tasks \cite{mazurowski2023segment, 10258199, zhang2023segment, raha2023generative, ren2024segment}. This specialization broadens models’ application versatility and enhances their performance in specialized scenarios.

Furthermore, the JFT-300M dataset \cite{jft-300m}, introduced by Google, is designed for training image classification models. It comprises 300 million images, each labelled by a sophisticated algorithm integrating raw web signals, interconnections between web pages, and user inputs. With over one billion labels and the possibility of multiple labels per image, JFT-300M offers a vast and diverse dataset for training robust models. Additionally, the LAION-5B \cite{laion5B} dataset provides a massive, openly accessible repository of image-text pairs, facilitating the training and improvement of large-scale multimodal models such as CLIP \cite{CLIP} and DALL-E \cite{DALL-E}.
Training models on more extensive and diverse datasets like ImageNet-21K, ImageNet-22K, and JFT-300M enables the development of more generalized and robust features. Consequently, models trained on these expansive datasets achieve higher performance on downstream tasks, particularly in scenarios that require generalization, such as FDG.

\subsection{Self Supervised Learning}\label{SSL}
Self-supervised learning (SSL) is an approach that allows deep learning models to learn from unlabeled data. This method is particularly valuable in scenarios where labeled data is scarce or prohibitively expensive to obtain. By leveraging pre-trained models trained with SSL algorithms, models can enhance accuracy by extracting additional insights from unlabeled data. Several key studies have introduced innovative frameworks in this domain.

Momentum Contrast (MoCo) \cite{he2020momentum} and its successor MoCov2 \cite{chen2020improved} are prominent frameworks designed for unsupervised visual representation learning. MoCo utilizes dynamic dictionaries for contrastive loss \cite{contrastiveLoss}, effectively ensuring the consistency of representations over time \cite{he2020momentum, chen2020improved}. MoCov2 enhances the original framework by improving augmentation strategies, adopting deeper network architectures, and implementing a cosine annealing learning rate schedule to boost learning efficiency and representation quality. Additionally, MoCov3 \cite{Chen_2021_ICCV} further improves training stability and efficiency by refining optimization techniques, leading to better performance on downstream tasks.

SwAV (Swapping Assignments between Views) \cite{caron2020unsupervised} introduces a unique approach where clusters of transformed images are swapped, promoting consistency and robust feature extraction without directly comparing image features. This method utilizes clustering techniques to manage assignments effectively, ensuring varied and comprehensive feature learning. On the other hand, DINO (Self-Distillation with No Labels) \cite{caron2021emerging} focuses on distilling knowledge within a network through self-distillation, where the outputs of one part of the network serve as targets for other parts. This internal feedback loop encourages the development of stable and discriminative features, facilitating learning without labelled data.

BEiT (BERT Pre-training of Image Transformers) \cite{bao2021beit} is inspired by successes in NLP, specifically utilizing the transformer architecture. It learns visual representations by predicting masked patches of an image, a process analogous to how BERT predicts missing words in a text sequence. This approach effectively adapts NLP techniques to visual data, enabling sophisticated pattern recognition without direct supervision. 

Pre-trained models developed using SSL are well-suited for Federated Domain Generalization (FDG). SSL enables models to learn rich, generalized feature representations from large-scale, unlabelled data, essential for adapting to diverse and unknown domains. In FDG, where data across clients varies significantly and labelled data is scarce, SSL pre-trained models can better capture underlying patterns. Therefore, exploring pre-trained models trained with state-of-the-art SSL techniques is valuable for enhancing FDG performance.

\subsection{Domain Generalization}\label{DG}
DG in deep learning involves developing models that can adapt to and perform accurately on new, previously unseen domains despite being trained only on data from specific source domains \cite{9847099, 27_muandet2013domain, 28_pan2018two, 29_seo2020learning, 6_zhou2021domain, 31_balaji2018metareg}. The objective is to create models that are robust to changes in data distribution, enabling them to handle real-world scenarios where training and testing data may significantly differ \cite{raha2024advancing, 9847099}. 
A predominant method for achieving DG is through domain-invariant representation learning, which seeks to minimize differences between source domains. The underlying assumption is that by creating a consistent representation across these domains, the model will generalize well to new domains. For instance, Muandet et al. \cite{27_muandet2013domain} aims to reduce domain disparities by employing Maximum Mean Discrepancy. Pan et al. \cite{28_pan2018two} integrate Instance Normalization and Batch Normalization in widely used CNNs, focusing on eliminating domain-specific features while preserving key discriminative aspects. In \cite{29_seo2020learning}, the authors report improved outcomes by incorporating these normalization techniques within an adjustable framework.

Data augmentation is another effective method for DG, enhancing the diversity of training data to better encompass the variability seen in test domains. Zhou et al. \cite{6_zhou2021domain} innovate by mixing training instance styles, which indirectly synthesizes new domains. The field has also seen increased interest in meta-learning approaches. Balaji et al. \cite{31_balaji2018metareg} propose MetaReg, which tailors a regularization function specifically for the network’s classification layer, separate from the feature extraction component. However, the effectiveness of these approaches typically relies on the availability of a diverse set of source domains, which may not be feasible in decentralized environments.

\subsection{Federated Learning}\label{FL}
FL \cite{2_qiao2023mp, 3_liu2021feddg,4_chen2023federated, 5_wu2021collaborative, 6_zhou2021domain} is an approach that allows several users to collaboratively train a deep learning model without requiring them to share their data. By design, FL protects user privacy, allowing participants to retain control over their own data. FedAvg \cite{15_fedavg} is an influential early work that catalyzed interest in FL among the deep learning community. Subsequently, numerous FL strategies have emerged, tackling challenges related to communication overhead \cite{16_reisizadeh2020fedpaq, 17_hamer2020fedboost}, data heterogeneity \cite{18_li2020federated, 19_karimireddy2020scaffold}, susceptibility to adversarial attacks \cite{20_wang2020attack, 21_park2021sageflow, 22_fang2020local}, and the need for model personalization \cite{23_deng2020adaptive, 24_li2021ditto, 25_fallah2020personalized}. Recently, Yuan et al. \cite{26_yuan2021we} identified two significant gaps in FL. Specifically, while existing approaches effectively manage out-of-sample variability within known client data distributions, they often fail to generalize to entirely unknown distributions. This limitation underscores the need for further research to enhance the adaptability and robustness of FL models in more diverse and unpredictable environments.

\subsection{Federated Domain Generalization}\label{FDG}
Although there is an increasing interest in tackling the DG problem inside the FL framework, especially in situations where each client has just one source domain, the current literature on this topic is still very limited. Liu et al. \cite{3_liu2021feddg} utilize the amplitude spectrum in the frequency domain to represent data distribution information, which they share among clients. However, these exchange operations can lead to increased costs and potential risks of data privacy breaches. Chen et al. \cite{4_chen2023federated} suggest a mechanism, CCST, that captures and shares the general style of local images across all clients. However, this method, like earlier approach, involves distributing some information about local data, potentially leading to data leakage, a critical concern in FL environments. Additionally, these techniques require the transmission and integration of partial data into new training sets, significantly increasing the communication and computational burdens throughout the FL training cycle.
Sun et al. \cite{33_sun2023feature} present FedKA, which incorporates a server-side voting mechanism to create pseudo-labels for the target domain. These labels are generated based on a consensus among clients, aiding in the fine-tuning of the global model.

In addressing domain shifts in FL, the work \cite{park2024stablefdg} introduces StableFDG, which incorporates DG through style-based learning and an attention-based feature highlighter. This approach enhances domain diversity and emphasizes critical features within classes, significantly outperforming existing DG baselines in data-poor FL scenarios.
In \cite{34_zhang2023federated}  Zhang et al. introduced a new global objective that includes a variance reduction regularizer to promote fairness. Following this, FedDG-GA\cite{34_zhang2023federated} was developed to optimize this objective through dynamic adjustment of aggregation weights. While these methods mitigate data leakage concerns, they also lead to significant resource consumption as the number of source domains and output classes grows.  In \cite{35_le2024efficiently}, the authors present gPerXAN, which employs a normalization approach guided by a specific regularizer. This architecture features Personalized Explicitly Assembled Normalization, designed to selectively filter out domain-specific features biased toward local data while preserving their discriminative attributes. Additionally, a simple but effective regularizer is introduced to direct these models towards capturing domain-invariant representations that enhance the efficacy of the global model’s classifier.

However, most previous studies in FDG have relied on using ResNet-18 or ResNet-50 models, pretrained on the ImageNet-1K dataset, as the backbone architecture for each client. Recently, more robust architectures like ViT, Swin Transformer and ConvNext have come to the fore. Additionally, models pretrained on larger datasets, such as ImageNet-22K \cite{36_russakovsky2015imagenet}, have potential to enhance performance in the FDG field. This evolution in model and dataset capabilities points to a research gap in exploring how these advanced architectures and larger datasets could further benefit FDG. The integration of these newer models and datasets may unlock higher levels of accuracy and generalizability, warranting deeper investigation.

\section{Methodology} \label{methodology}
This section begins by introducing the FL algorithm, detailing its collaborative training process and mathematical foundations. It then formulates the problem of FDG, outlining the challenges of building models that perform well across diverse and unseen domains. Finally, it provides an overview of our approach, describing the model adoption process, including the selection of architectures, pre-training datasets, implementation of self-supervised learning techniques, and the fine-tuning procedures tailored for this study.
\subsection{Preliminaries}\label{AA}
\begin{figure}[!t]
	\centerline{\includegraphics[width=9cm]{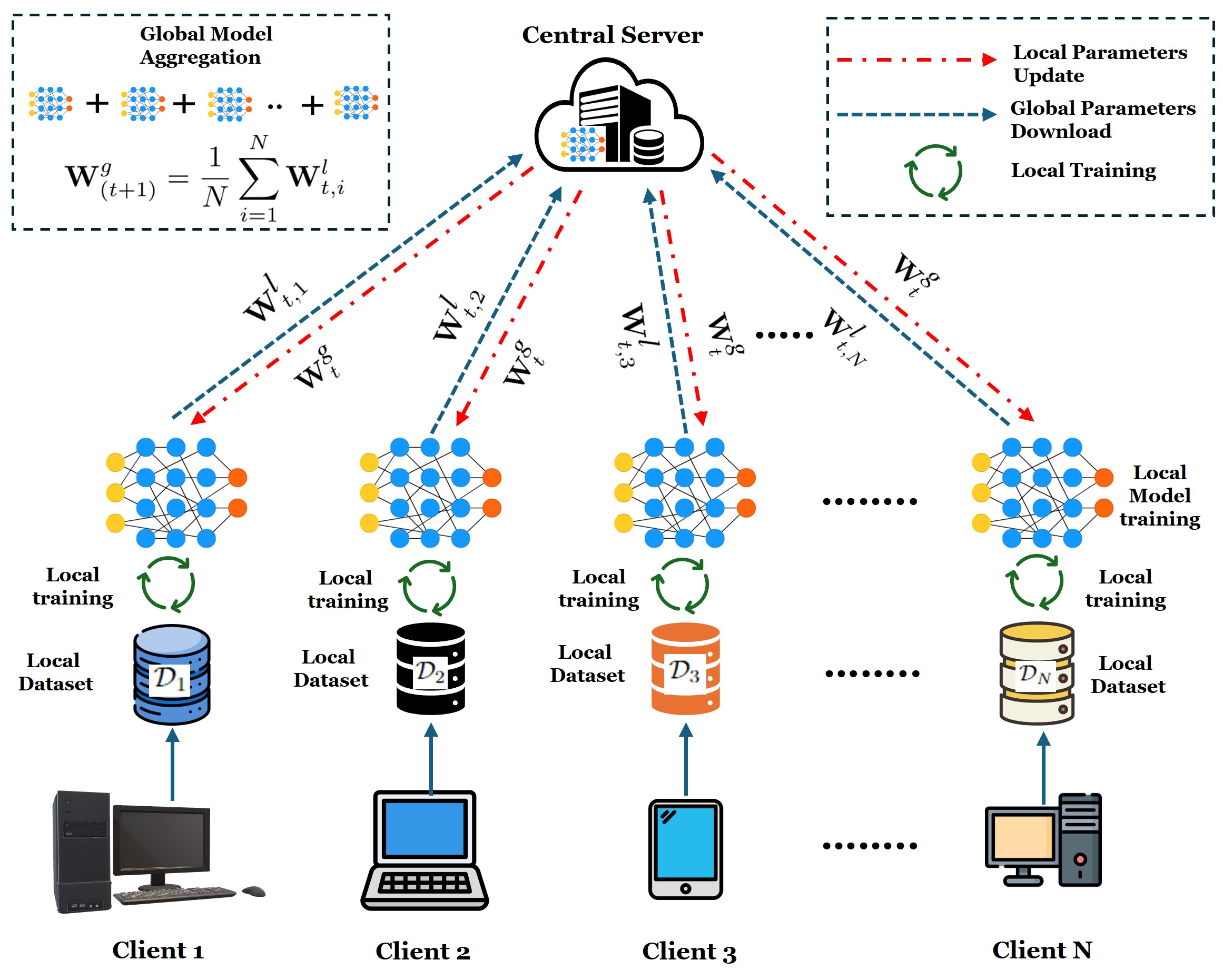}}
	\caption{Overview of the federated training process.}
	\label{System_model}
\end{figure}

FL is a technique that enables the collaborative training of a machine learning model by decentralized clients, each utilizing their own local data \cite{2_qiao2023mp,3_liu2021feddg,4_chen2023federated,5_wu2021collaborative}. This method is specifically designed to enhance data privacy by ensuring that data remains on the local devices of each client. The FL framework is illustrated in Fig. \ref{System_model}. The training process can be described as follows:

In the conventional FL procedure, each client \( i \) possesses a private and heterogeneous dataset, denoted as \( \mathcal{D}_i = \{(\xi_i^j, y_i^j)\}_{j=1}^{|\mathcal{D}_i|} \), where \( |\mathcal{D}_i| \) is the number of data points for client \( i \). Here, \( \xi_i^j \) represents the input features, and \( y_i^j \in \{1, \ldots, \mathcal{C}\} \) is the corresponding label for the \( j \)-th data point, with \( \mathcal{C} \) being the total number of classes. The objective is to coordinate the collaboration between clients and the edge server to collectively train a shared model \( \mathcal{F}(\mathbf{W}^g; \xi_i) \). The empirical risk for client \(i\), typically exemplified by cross-entropy loss with one-hot encoded labels, is defined as follows \cite{2_qiao2023mp}:
\vspace{-3mm}
\begin{equation}
f_i(\mathbf{W}^g) = -\sum_{j=1}^{\mathcal{C}} \mathbf{1}_{(y_i=j)} \log p_j(\mathcal{F}(\mathbf{W}^g; \xi_i); y_i).
\label{eq:empirical-risk}
\end{equation}
\vspace{-1mm}
where \( \mathbf{1}_{(\cdot)} \) represents the indicator function, \( \mathbf{W}^g \) represents the shared parameters of the global model, \( \mathcal{C} \) represents the number of classes in the label space \( [\mathcal{C}] = \{1, \ldots, \mathcal{C}\} \), and \( p_j(\mathcal{F}(\mathbf{W}^g; \xi_i) ; y_i) \) represents the probability of the data sample \( \xi_i ;y_i \) being classified as the \( j \)-th class.
Furthermore, the training process for each individual client aims to minimize the local loss specific to that client. The local loss \( L_i(\mathbf{W}^g) \) is defined as follows \cite{2_qiao2023mp}:
\vspace{-3mm}
\begin{equation}
L_i(\mathbf{W}^g) = \frac{1}{|\mathcal{D}_i|} \sum_{(\xi_i^j, y_i^j) \in \mathcal{D}_i} f_i(\mathbf{W}^g).
\label{eq:local-loss}
\end{equation}
The global objective is to minimize the weighted sum of the local losses across all clients as follows:
\begin{equation}
\arg \min_{\mathbf{W}^g} L(\mathbf{W}^g) = \sum_{i \in [N]} \frac{|\mathcal{D}_i|}{\sum_{k \in [N]} |\mathcal{D}_k|} L_i(\mathbf{W}^g).
\label{eq:global-loss}
\end{equation}
where \( [N] \) denotes the set of distributed clients, \( [N] = \{1, \ldots, N\} \).

\begin{figure*}[!t]
	\centerline{\includegraphics[width=15cm]{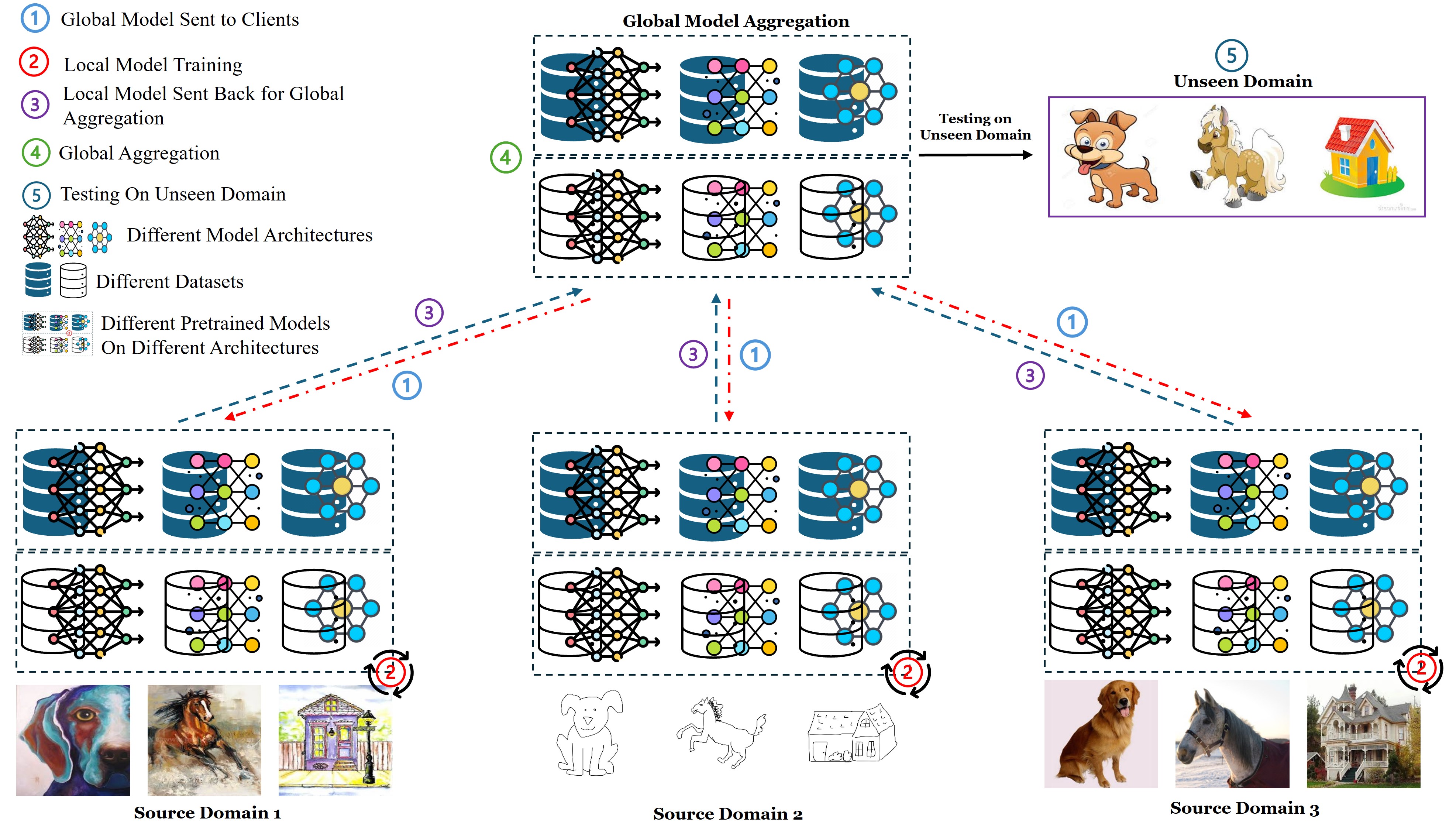}}
	\caption{Federated Domain Generalization (FDG) Framework: Leveraging Pre-trained Models for Cross-Domain Adaptation. This illustration captures the core of our FDG strategy, where pre-trained SOTA models are distributed to clients for local adaptation. Each client fine-tunes the model with local data, while the central server aggregates updates to build a robust global model, enabling powerful domain generalization across diverse environments.}
	\label{overview}
\end{figure*}

\subsection{Problem Formulation}

Let \( \mathbf{E} \) and \( \mathbf{Y} \) represent the input and label spaces, respectively, for a given task \( T \). In a typical FL scenario, as discussed in an earlier subsection, we consider \( N \) clients, each denoted by \( c_i \) for \( i = 1, \ldots, N \). Each client possesses a dataset \( \mathcal{D}_i = \{(\xi_k, y_k)\}_{k=1}^{|\mathcal{D}_i|} \), characterized by a distinct domain governed by a joint probability distribution \( \mathbf{P}^{(i)}_{\mathbf{E},\mathbf{Y}} \). It is important to note that the datasets across different clients are diverse and satisfy the following conditions \cite{35_le2024efficiently}:
\begin{itemize}
    \item \( \mathbf{P}^{(i)}_{\mathbf{E},\mathbf{Y}} \neq \mathbf{P}^{(j)}_{\mathbf{E},\mathbf{Y}} \) for all \( i, j \) where \( i \neq j \), indicating the heterogeneity in the data distribution across clients.
    \item \( \mathbf{P}^{(i)}_{\mathbf{Y}|\mathbf{E}} = \mathbf{P}^{(j)}_{\mathbf{Y}|\mathbf{E}} \) for all \( i, j \) where \( i \neq j \), suggesting that despite differences in joint distributions, the conditional distributions of the outputs given inputs are consistent across different domains.
\end{itemize}
The goal of FDG is to utilize these diverse datasets to construct a global model, \( \mathcal{F}(\mathbf{W}^g; \xi_i) \), that is capable of generalizing to an unseen dataset \( \mathcal{D}_U \) from an unseen domain. This global model should satisfy the following conditions \cite{35_le2024efficiently}:
\begin{itemize}
    \item \( \mathbf{P}^{(U)}_{\mathbf{E},\mathbf{Y}} \neq \mathbf{P}^{(i)}_{\mathbf{E},\mathbf{Y}} \) for all \( i \) from 1 to \( N \), indicating that the joint distribution of inputs and labels in the unseen domain differs from that of any training domain.
    \item \( \mathbf{P}^{(U)}_{\mathbf{Y}|\mathbf{E}} = \mathbf{P}^{(i)}_{\mathbf{Y}|\mathbf{E}} \) for all \( i \) from 1 to \( N \), suggesting that the conditional distribution of labels given inputs remains consistent across both seen and unseen domains.
\end{itemize}
To achieve this, the \( N \) clients collaborate with a central server over \( T \) rounds. In each round, every client \( c_i \) receives the same initial global model \( \mathcal{F}(\mathbf{W}^g; \xi_i) \) from the server and updates it using their local dataset \( \mathcal{D}_i \) over \( K \) epochs to form their local model \( f_i \). The server then aggregates these local models to update the global model. This aggregation process continues until the global model converges.

\subsection{Overview of the approach} 
In this study, we explore various SOTA architectures that have been pre-trained on a range of benchmark datasets. The primary goal is to evaluate the effectiveness of these SOTA models in the context of FDG. Figure 2 visually illustrates the conceptual framework underpinning our research inquiry into FDG. In our setup, each client $c_i$ consists of a base model $\mathcal{M}_{i}^{\text{base}}$. 

\textbf{Base Model Configuration:}
The base model at each client $c_i$ includes a feature extractor and a classifier, which are initially configured as follows:
\begin{equation}
\mathcal{G}^{\text{features}}_{c_i} = \Phi^{\text{extractor}}_{c_i}(x_{\mathcal{D}_i}; \theta^{\text{parameters}}_{c_i}), \label{eq:4}
\end{equation}
\vspace{-5mm}
\begin{equation}
\mathcal{J}^{\text{base}}_{c_i} = \Psi^{\text{classifier}}_{c_i}(\mathcal{G}^{\text{features}}_{c_i}; \phi^{\text{parameters}}_{c_i}), \label{eq:5}
\end{equation}
\vspace{-5mm}
\begin{equation}
\mathcal{M}_{i}^{\text{base}}(x) = \mathcal{J}^{\text{base}}_{c_i}. \label{eq:6}
\end{equation}

Where $\Phi^{\text{extractor}}_{c_i}$ is the feature extractor of the client $c_i$, the classifier of the client $c_i$ is denoted as $\Psi^{\text{classifier}}$, $x_{\mathcal{D}_i}$ is input to the feature extractor $\Phi^{\text{extractor}}_{c_i}$, the output of the feature extractor $\Phi^{\text{extractor}}_{c_i}$, is denoted as $\mathcal{G}^{\text{features}}_{c_i}$, $\theta^{\text{parameters}}_{c_i}$ and $\phi^{\text{parameters}}_{c_i}$ are the parameters of the feature extractor and classifiers respectively.

\textbf{Model Adaptation:}
The classifier of the base model is replaced with new classifiers suited to the local dataset $\mathcal{D}_i$.
\begin{equation}
\mathcal{M}_{i,\text{new}}^{\text{base}}(x) = \Psi^{\text{classifier}}_{{c_i},\text{new}}(\Phi^{\text{extractor}}_{c_i}(x_{\mathcal{D}_i}; \theta^{\text{parameters}}_{c_i}); \phi^{\text{parameters}}_{{c_i},\text{new}}). \label{eq:7}
\end{equation}
At the beginning of the process, the central server sends the global model parameters $\mathbf{W}^g$ to each FL client \(c_i\). The global model parameters $\mathbf{W}^g$ can be represented as,
\begin{equation}
\mathbf{W}^g = (\theta^g; \phi^g).
\label{eq:8}
\end{equation}
Here, $\theta^g$ is the weights obtained from training benchmark datasets and $\phi^g$ is configured for initial classifier settings. The weight initialization for each client $c_i$ can be written as:
\begin{equation}
\theta^{\text{parameters}}_{c_i} = \theta^g.
\end{equation}
\vspace{-5mm}
\begin{equation}
\phi^{\text{parameters}}_{{c_i},\text{new}} = \phi^g.
\end{equation}
Each client starts with these pre-trained global model parameters $\mathbf{W}^g$ and trains its model $\mathcal{M}_i$ with their local data dataset $\mathcal{D}_i$. 

\textbf{Fine-tuning:}
The next step is to fine-tune the feature extractor and the newly adapted classifier using the local datasets:
\begin{equation}
\begin{aligned}
&(\theta_{{c_i},{\text{fine-tune}}}^{\text{parameters}}, \phi_{{c_i},{\text{fine-tune}}}^{\text{parameters}}) \\
&= \arg\min_{\theta_{{c_i},{\text{fine-tune}}}^{\text{parameters}}, \phi_{{c_i},{\text{fine-tune}}}^{\text{parameters}}} 
\left[ \mathcal{L}\left(\mathcal{M}_{i,\text{new}}^{\text{base}}(\xi_k), y_k\right) \right]
\end{aligned}
\label{eq:6}
\end{equation}

where \(\mathcal{L}(.)\) denotes the loss function which can be represented as:
\begin{equation}
\mathcal{L}(\mathcal{M}_{i,\text{new}}^{\text{base}}(\xi), y) = -\sum_{c=1}^C y_c \log(\mathcal{M}_{i,\text{new}}^{\text{base}}(\xi_c)),
\end{equation}
where \( C \) is the number of classes, \( y_c \) is the true probability of class \( c \) (usually 0 or 1 in one-hot encoding), and \( \mathcal{M}_{i,\text{new}}^{\text{base}}(\xi_c) \) is the predicted probability of class \( c \) from the model.
Each model $c_i$ trains their model $\mathcal{M}_i$ with the local dataset $\mathcal{D}_i$ for $\mathcal{N}$ iterations and sends the updated local parameters $\mathbf{W}^l_i$ to the central server for the global weight aggregation. The central server receives the local models and then aggregates them by using this formula: 
\begin{equation}
\mathbf{W}^{g}_{\text{new}} = \frac{1}{\sum_{i=1}^N n_i} \sum_{i=1}^N n_i \mathbf{W}^l_i,
\end{equation}
where $N$ is the total number of clients participating in the training, $n_i$ is the number of data samples from the $i$-th client, $\mathbf{W}^l_i$ represents the local model parameters from the $i$-th client after local training and $\mathbf{W}^{g}_{\text{new}}$ is the new set of global model parameters obtained after aggregating the updates.
After the aggregation, the  server transmits the updated global model parameters $\mathbf{W}^g$ to the clients for the next round and this fine-tuning process continues until convergence.

\subsubsection{Network Architectures and Pre-training Datasets} 
We utilized a diverse array of pre-trained models for our network architectures, encompassing several variants of ConvNeXt \cite{ConvNeXt}, including ConvNeXt Base (ConvNeXt-B), ConvNeXt Small (ConvNeXt-S), ConvNeXt Tiny (ConvNeXt-T), ConvNeXt Nano (ConvNeXt-N), ConvNeXt Atto (ConvNeXt-A), and ConvNeXt Pico (ConvNeXt-P). Additionally, we incorporated various ResNet architectures such as ResNet-18, ResNet-50 \cite{12_he2016deep}, and the enhanced ResNetv2 \cite{resNetv2}. Our exploration extended to Transformer-based models, where we employed Swin Transformers (Swin-B, Swin-S, Swin-T) \cite{13_liu2021swin} alongside Vision Transformers (ViT-B, ViT-S) \cite{dosovitskiy2020image}.

Due to the computational constraints inherent to edge devices within federated setups, we excluded larger model variants such as ConvNeXt Large, ViT Huge, and ResNet-101, concentrating instead on models optimized for decentralized environments. The pre-training datasets leveraged in our study included ImageNet-1K \cite{36_russakovsky2015imagenet}, ImageNet-21K, ImageNet-12K, ImageNet-22K, and JFT-300M \cite{jft-300m}. Training on these expansive and diverse datasets facilitated robust feature learning and enhanced model generalization, thereby supporting effective Federated Domain Generalization (FDG). Although other datasets like COCO and SA-1B are available, we did not incorporate them due to the unavailability of pre-trained models or their limited relevance to our specific FDG applications.

\begin{table*}[h]
\centering
\caption{Performance metrics for different models with parameter sizes pretrained on ImageNet-1k and evaluated on the Office-Home dataset.}
\label{Tab1}
\begin{tabular}{lcccccccc}
\toprule
\textbf{Model} & \textbf{Parameters} & \textbf{Pretrained Dataset}  & \textbf{P} & \textbf{C} & \textbf{A} & \textbf{R} & \textbf{Avg} \\
\midrule
ResNet-18 & 11.93M & ImageNet-1K  & 0.700834 & 0.466438 & 0.515039
 & 0.702089 & 0.596099 \\
ResNet-50 & 25.77M & ImageNet-1k  & 0.803334 & 0.548225 & 0.668315 & 0.810420 & 0.707573 \\
ResNetv2-101 & 44.75M & ImageNet-1k  & 0.798829 & 0.610767 & 0.737124 & 0.823502 & 0.742555 \\
ViT-S & 22.65M & ImageNet-1k  & 0.752647 & 0.512715 & 0.637412 & 0.777140 & 0.669978 \\
ViT-B & 87.07M & ImageNet-1k & 0.784411 & 0.585796 & 0.678204 & 0.811797 & 0.715052 \\
Swin-T & 28.79M & ImageNet-1k  & 0.808515 & 0.548912 & 0.686032 & 0.823043 & 0.716625 \\
Swin-S & 50.12M & ImageNet-1k  & 0.840955 & 0.605269 & 0.742892 & 0.851503 & 0.761848 \\
Swin-B & 88.22M & ImageNet-1k  & 0.849290 & 0.596334 & 0.748661 & 0.853110 & 0.761849 \\
ConvNeXt-A & 4.31M & ImageNet-1k & 0.757378 & 0.508591 & 0.602390 & 0.763369 & 0.657932\\
ConvNeXt-P & 9.61M & ImageNet-1k  & 0.781257 & 0.558763 & 0.655130 & 0.796190 & 0.697835\\
ConvNeXt-N & 16.13M & ImageNet-1k  & 0.815950 & 0.565636 & 0.697981 & 0.820289 & 0.724964\\
ConvNeXt-T & 29.10M & ImageNet-1k  & 0.838027 & 0.597251 & 0.739184 & 0.846454 & 0.755229\\
ConvNeXt-S & 50.15M & ImageNet-1k  & \textbf{0.868214} & 0.613975 & 0.761434 & 0.860225 & 0.776000\\
ConvNeXt-B & 89.05M & ImageNet-1k  & 0.865060 & \textbf{0.621764} & \textbf{0.788216} & \textbf{0.861602} & \textbf{0.784161} \\
\bottomrule
\end{tabular}
\end{table*}

\begin{table*}[h]
\centering
\caption{Performance metrics for different models with parameter sizes pretrained on various datasets and evaluated on the Office-Home dataset.}
\label{tab2}
\begin{tabular}{lcccccccc}
\toprule
\textbf{Model} & \textbf{Parameter Size} & \textbf{Pretrained Dataset} & \textbf{P} & \textbf{C} & \textbf{A} & \textbf{R} & \textbf{Avg} \\
\midrule
ResNetv2-101 & 44.75M & JFT-300M  & 0.759180 & 0.556243 & 0.641121 & 0.791600 & 0.687036 \\
ViT-S & 22.65M & ImageNet-21K  & 0.835774 & 0.619015 & 0.743717 & 0.843011 & 0.760379 \\
ViT-B & 87.07M & ImageNet-21K  & 0.801532 & 0.627262 & 0.711166 & 0.811338 & 0.737824 \\
Swin-T & 28.79M & ImageNet-22K  & 0.850642 & 0.617182 & 0.748249 & 0.853110 & 0.767295 \\
Swin-S & 50.11M & ImageNet-22K  & 0.881730 & 0.682932 & 0.803049 & 0.886390 & 0.813525 \\
Swin-B & 88.22M & ImageNet-22K  & 0.878802 & 0.687285 & 0.800165 & 0.881799 & 0.812012 \\
ConvNeXt-N & 16.13M & ImageNet-12K  & 0.815724 & 0.601833 & 0.714050 & 0.819142 & 0.737687 \\
ConvNeXt-T & 29.10M & ImageNet-22K  & 0.879928 & 0.667354 & 0.785744 & 0.875602 & 0.802157 \\
ConvNeXt-S & 50.14M & ImageNet-22K  & 0.889615 & 0.697365 & 0.816646 & 0.897177 & 0.825201 \\
ConvNeXt-B & 50.14M & ImageNet-22K  & \textbf{0.898175} & \textbf{0.733104} & \textbf{0.844252} & \textbf{0.903144} & \textbf{0.844668} \\
\bottomrule
\end{tabular}
\end{table*}

\begin{table*}[h]
\centering
\caption{Performance metrics for different self-supervised learning mechanisms using ResNet-50 architecture pretrained on ImageNet-1k and evaluated on the Office-Home dataset.}
\label{tab3}
\begin{tabular}{lccccccccc}
\toprule
\textbf{Mechanism} & \textbf{Architecture} & \textbf{Parameters} & \textbf{Pretrained Dataset} &  \textbf{P} & \textbf{C} & \textbf{A} & \textbf{R} & \textbf{Avg} \\
\midrule
MoCov3 & ResNet-50 & 26.68M & ImageNet-1K  & 0.753098 & 0.529210 & 0.618047 & 0.763599 & 0.665988 \\
SwAV & ResNet-50 & 26.68M & ImageNet-1K  & 0.759405 & 0.479725 & 0.611867 & 0.768878 & 0.654968 \\
DINO & ViT-S & 21.86M & ImageNet-1K  & 0.163325 & 0.104467 & 0.105892 & 0.175809 & 0.137373 \\
BEiT & ViT-B & 87.04M & ImageNet-21K  & \textbf{0.833296} & \textbf{0.676060} & \textbf{0.763494} & \textbf{0.851962} & \textbf{0.781203} \\
\bottomrule
\end{tabular}
\end{table*}

\begin{figure}[!t]
	\centerline{\includegraphics[width=9cm]{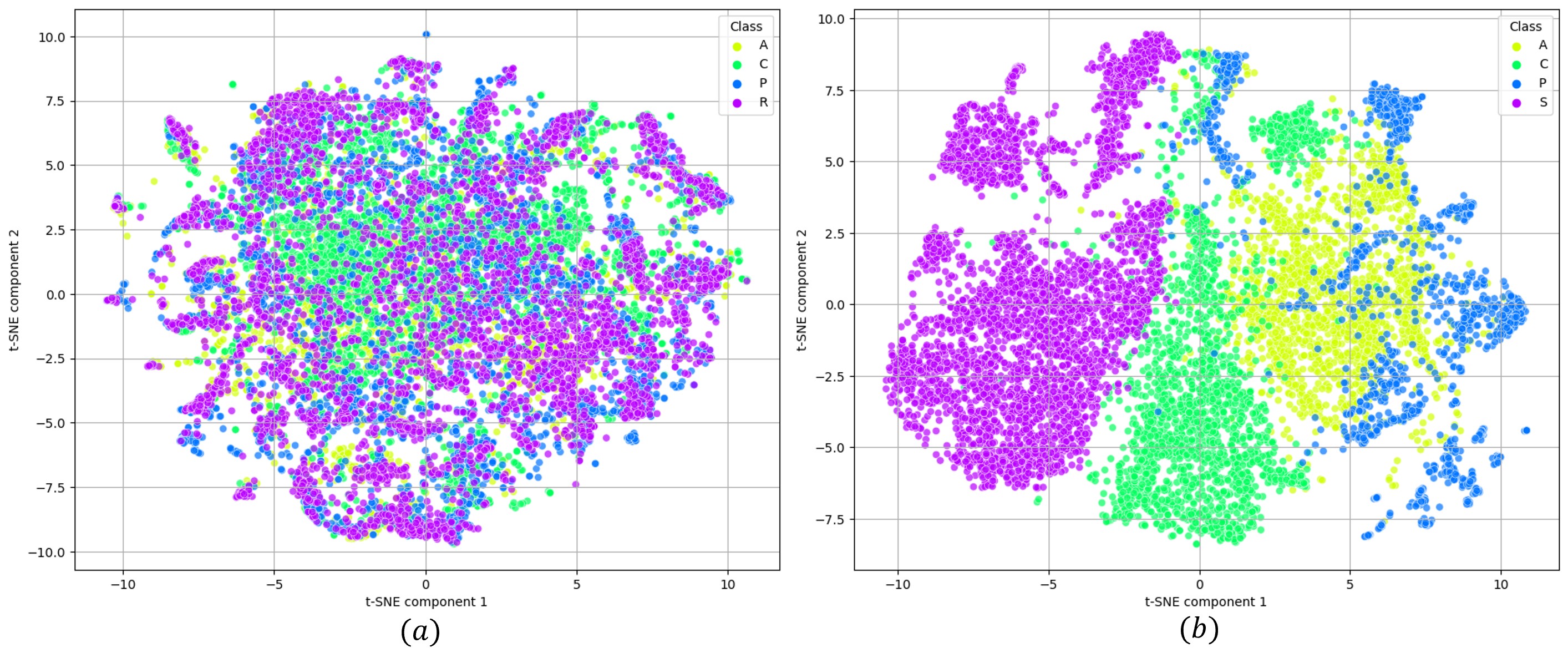}}
	\caption{t-SNE visualization of feature embeddings extracted from images (a) in the Office-Home dataset and (b) in the PACS dataset using a pre-trained MobileNetV2 model. The plot represents the distribution of image features across different classes after dimensionality reduction to two components, highlighting the separability and clustering of data points corresponding to different domains.}
	\label{tnse_domains}
\end{figure}
\subsubsection{Self-Supervised Learning Techniques}
In addition to supervised pre-training, we investigated several pretrained models which are trained with SSL techniques. Our study encompassed MoCov3 \cite{chen2020improved}, DINO \cite{Chen_2021_ICCV}, SwAV \cite{caron2021emerging}, and BEiT \cite{bao2021beit}. These SSL techniques were selected for their exceptional ability to leverage unlabeled data, thereby deriving robust and generalizable feature representations essential for achieving success across varied and unseen domains characteristic of federated environments.

\subsubsection{Downstream Datasets} 
We conduct experiments on two widely recognized benchmark datasets for FDG, namely PACS \cite{li2017deeper} and Office-Home \cite{venkateswara2017deep}. PACS consists of four domains, each exhibiting significant variations in image colors and textures: Photo (P), Art painting (A), Cartoon (C), and Sketch (S). Each domain encompasses seven categories, totaling 9,991 images. The domains in PACS are well distinguishable, as the variations in colors, textures, and artistic styles result in clearly separable clusters in feature space. Similarly, Office-Home comprises four domains characterized by smaller discrepancies in backgrounds and camera viewpoints: Product (P), Art (A), Clipart (C), and Real-world (R). Each domain features a more extensive label set, comprising 65 categories with a total of 15,588 images. Fig \ref{tnse_domains} shows t-SNE visualization of feature embeddings extracted from images (a) in the Office-Home dataset and (b) in the PACS dataset using a pre-trained MobileNetv2 model. The plot represents the distribution of image features across different domains after dimensionality reduction to two components, highlighting the separability and clustering of data points corresponding to different domains. From the figure we can see that the domains in Office-Home exhibit significant overlap in the feature space.  In contrast, the domains in the PACS dataset are clearly distinct, reflecting the more pronounced differences in data distribution.

\subsubsection{Finetuning}
After obtaining pre-trained models, we fine-tune them using downstream datasets. In our experiments, we use a fixed learning rate of 0.0001 for client fine-tuning. Additionally, we utilize a new fully connected (FC) layer designed for the downstream tasks. Then each client model is fine-tuned using the Adam optimizer, which is known for its efficiency in handling sparse gradients and adaptive learning rate capabilities. The update rule for the Adam optimizer is defined as follows:
\begin{align}
m_t &= \beta_1 m_{t-1} + (1 - \beta_1) g_t, \\
v_t &= \beta_2 v_{t-1} + (1 - \beta_2) g_t^2, \\
\hat{m}_t &= \frac{m_t}{1 - \beta_1^t}, \\
\hat{v}_t &= \frac{v_t}{1 - \beta_2^t}, \\
\theta_{t+1} &= \theta_t - \frac{\eta \hat{m}_t}{\sqrt{\hat{v}_t} + \epsilon}.
\end{align}
where $m_t$ and $v_t$ are estimates of the first moment (the mean) and the second moment (the uncentered variance) of the gradients respectively, $g_t$ is the gradient at time step $t$, $\beta_1$ and $\beta_2$ are decay rates which control the moving averages of the gradient and its square, $\eta$ is the learning rate, and $\epsilon$ is a small scalar used to prevent division by zero.
To ensure consistency across experiments, we maintain a fixed image size of 224 × 224 with random resized cropping. Moreover, we incorporate data augmentation techniques such as random color jittering, gray-scaling, and horizontal flipping to enrich the training data and improve model robustness.

\section{Experimental setup and Numerical Results} \label{experiment}
In this section, we first outline the experimental setup of the study, followed by a presentation of the results on both datasets. Next, we discuss the impact of self-supervised and supervised pretraining. We then explore the effects of different architectures, pretraining datasets, and parameters.

\begin{figure*}[!t]
	\centerline{\includegraphics[width=16cm]{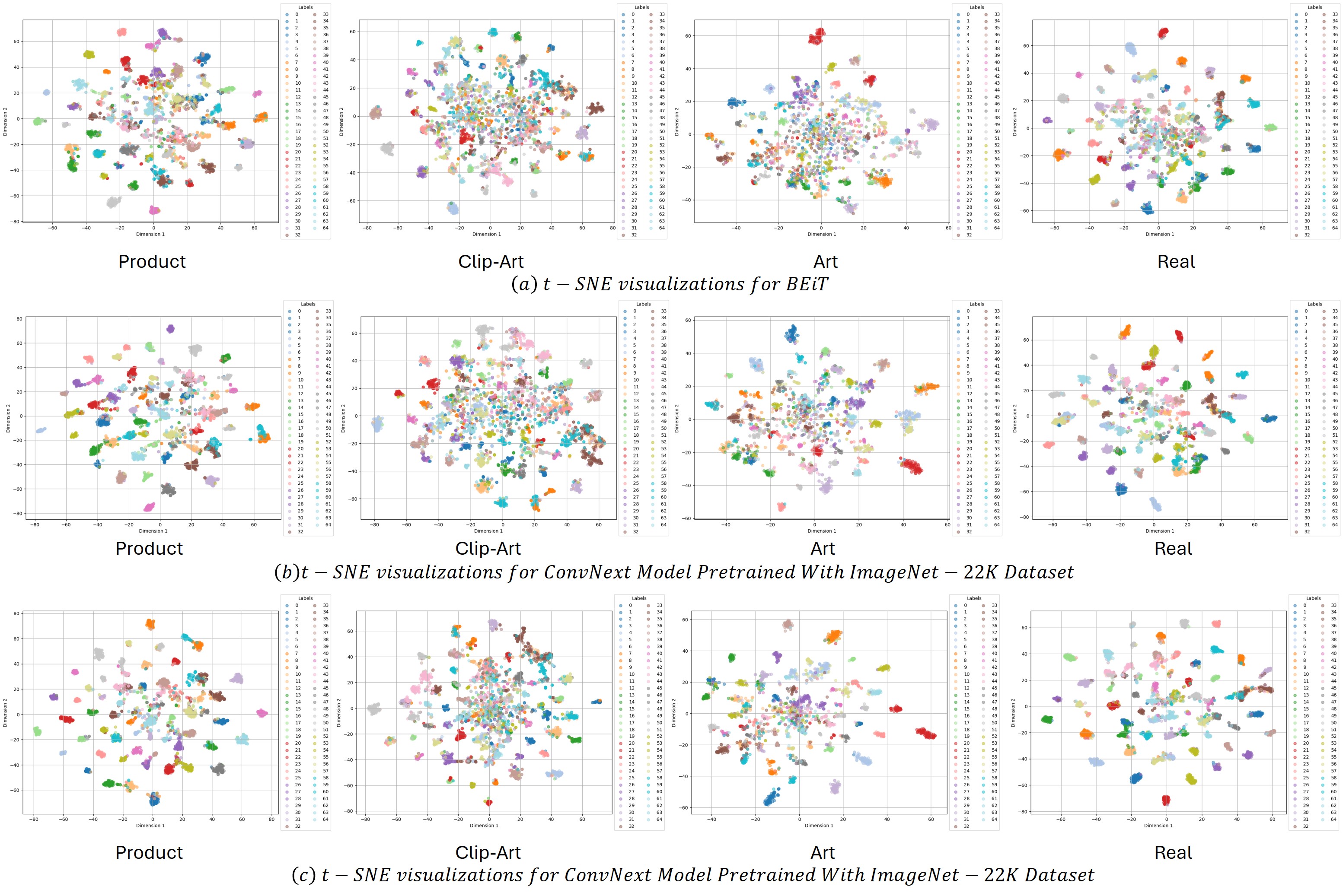}}
	\caption{t-SNE plots for feature representations from different models: BEiT pretraining, ConvNeXt-B pretrained on ImageNet-1K, and ConvNeXt-B pretrained on ImageNet-22K for the Office-Home dataset. The ConvNeXt-B model pretrained on ImageNet-22K demonstrates better clustering and distinctiveness of features compared to the other models, suggesting a more effective feature space adaptation, which likely contributes to its superior performance in DG tasks.}
	\label{tnse01}
\end{figure*}
\subsection{Environment Setup and Data Partition}
The research is conducted using a sophisticated configuration on a desktop computer equipped with a Core i9 processor (2.8 GHz). We utilized the Pytorch version 2.0.1 and Python version 3.11.4, running on a 64-bit operating system. The system is equipped with a powerful Geforce RTX 3080ti GPU. For a fair comparison, we follow the widely used leave-one-domain-out evaluation protocol used in \cite{3_liu2021feddg}, \cite{4_chen2023federated} , \cite{34_zhang2023federated}, \cite{35_le2024efficiently}. Specifically, we iteratively select one domain as the unseen domain, train the model on the remaining domains (with each domain treated as a separate client), and then evaluate the trained model on the selected unseen domain. 
In each communication round, the client model was trained for five epochs. Subsequently, these client models were aggregated to construct the global model by using the FedAvg \cite{15_fedavg} algorithm. This process was repeated across $20$ communication rounds.

\subsection{Results on Office-Home dataset}
In this section, we present a comprehensive analysis of various deep learning models evaluated on the Office-Home dataset, focusing on their domain generalization capabilities. The models include ResNet variants, Vision Transformers (ViT), Swin Transformers, ConvNeXt models, and several SSL methods.

Table \ref{Tab1} showcases the performance of models pretrained on the ImageNet-1K dataset. The ResNet series demonstrates consistent improvements with increasing depth: ResNet-18 achieves a mean accuracy of 59.61\%, ResNet-50 improves to 70.76\%, and ResNetv2-101 reaches 74.26\%. ViT models also exhibit strong performance, with ViT-B achieving a mean accuracy of 71.51\%. Swin Transformers further enhance the results, with both Swin-S and Swin-B attaining average accuracies of 76.18\%. However, the smaller Swin-T model shows a performance drop of 4.52\% compared to its larger counterparts.

The ConvNeXt models, specifically ConvNeXt-S and ConvNeXt-B, dominated the performance rankings, obtaining the highest accuracies. ConvNeXt-B achieved a zenith of 78.41\%, while ConvNeXt-S achieved an impressive 77.60\% average accuracy. The smaller versions of the ConvNeXt models (ConvNeXt-P, ConvNeXt-N, ConvNeXt-A) showed performance drops compared to their larger counterparts. However, they still maintained commendable performance, with ConvNeXt-P reaching 69.78\%, ConvNeXt-N 72.50\%, and ConvNeXt-A 65.79\% on average accuracy. This demonstrates the strong scaling capabilities of the ConvNeXt architecture, as even the smaller models perform competitively in domain generalization tasks.  
\begin{table*}[h]
\centering
\caption{Performance metrics for different models with parameter sizes pretrained on ImageNet-1k and evaluated on the PACS dataset.}
\label{tab4}
\begin{tabular}{lccccccccc}
\toprule
\textbf{Model} & \textbf{Parameters} & \textbf{Pretrained Dataset}  & \textbf{P} & \textbf{A} & \textbf{C} & \textbf{S} & \textbf{Avg} \\
\midrule
ResNet-18 & 11.93M & ImageNet-1K  & 0.924551 & 0.731445 & 0.662116 & 0.609061 & 0.731293 \\
ResNet-50 & 25.77M & ImageNet-1k  & 0.952096 & 0.786621 & 0.723976 & 0.703487 & 0.791545 \\
ResNetv2-101 & 44.75M & ImageNet-1k  & 0.981437 & 0.838867 & 0.756826 & 0.739374 & 0.829126 \\
ViT-S & 22.65M & ImageNet-1k  & 0.970060 & 0.816406 & 0.770051 & 0.687961 & 0.811119 \\
ViT-B & 87.07M & ImageNet-1k  & 0.974251 & 0.839355 & 0.770051 & 0.720285 & 0.825985 \\
Swin-T & 28.79M & ImageNet-1k  & 0.983832 &  0.875000 & 0.791809 & 0.753118 & 0.850940 \\
Swin-S & 50.11M & ImageNet-1k  & 0.986826 & 0.894043 & 0.821672 & 0.771952 & 0.868623 \\
Swin-B & 88.22M & ImageNet-1k  & 0.988623 & 0.902344 & 0.827218 & 0.761517 & 0.869926 \\
ConvNeXt-A & 4.31M & ImageNet-1k  & 0.953293 & 0.839844 & 0.784556 & 0.766353
 & 0.836011 \\
ConvNeXt-P & 9.61M  & ImageNet-1k  & 0.969461 & 0.862793 & 0.813140 & 0.799695
 & 0.861272 \\
ConvNeXt-N & 16.13M & ImageNet-1k  & 0.983234 & 0.892090 & 0.819966 & 0.783914 & 0.869801 \\
ConvNeXt-T & 29.10M & ImageNet-1k  & 0.992216 & 0.919922 & 0.824232 & 0.784678 & 0.880012 \\
ConvNeXt-S & 50.14M & ImageNet-1k  & 0.989222 & 0.905273 & 0.849829 & \textbf{0.848816} & 0.898284 \\
ConvNeXt-B & 89.04M & ImageNet-1k & \textbf{0.992216} & \textbf{0.933105} & \textbf{0.868174} & 0.835582 & \textbf{0.907269} \\
\bottomrule
\end{tabular}
\end{table*}

\begin{table*}[h]
\centering
\caption{Performance metrics for different models with parameter sizes pretrained on various datasets and evaluated on the PACS dataset.}
\label{tab5}
\begin{tabular}{lccccccccc}
\toprule
\textbf{Model} & \textbf{Parameters} & \textbf{Pretrained Dataset} & \textbf{P} & \textbf{A} & \textbf{C} & \textbf{S} & \textbf{Avg} \\
\midrule
ResNetv2-101 & 44.75M & JFT-300M  & 0.961078 & 0.801758 & 0.814846 & 0.781878 & 0.839890 \\
ViT-S & 22.65M & ImageNet-21k  & 0.987425 & 0.879883 & 0.805461 & 0.763808 & 0.859144 \\
ViT-B & 87.07M & ImageNet-21k  & 0.952096 & 0.814941 & 0.758532 & 0.786460 & 0.828007 \\
Swin-T & 28.79M & ImageNet-22K  & 0.991018 & 0.895508 & 0.824232 & 0.787987 & 0.874686 \\
Swin-S & 50.11M & ImageNet-22K  & 0.995210 & 0.932617 & 0.847270 & 0.843726 & 0.904705 \\
Swin-B & 88.22M & ImageNet-22K  & 0.995210 & 0.932617 & 0.853669 & 0.856452 & 0.909487 \\
ConvNeXt-N & 16.13M & ImageNet-12K  & 0.976647 & 0.860352 & 0.821246 & 0.825146 & 0.870847 \\
ConvNeXt-T & 29.10M & ImageNet-22K  & 0.992814 & 0.922852
 & 0.857935 & 0.822856 & 0.899114 \\
ConvNeXt-S & 50.14M & ImageNet-22K & \textbf{0.997857} & 0.949707 & \textbf{0.894198} & 0.847544 & \textbf{0.922326} \\
ConvNeXt-B & 89.04M & ImageNet-22K  & 0.997605 & \textbf{0.956055} & 0.880973 & \textbf{0.867396} & 0.925507 \\
\bottomrule
\end{tabular}
\end{table*}

\begin{table*}[h]
\centering
\footnotesize
\caption{Performance metrics for different self-supervised mechanisms and architectures with parameter sizes pretrained on various datasets and evaluated on the PACS dataset.}
\label{tab6}
\begin{tabular}{lccccccccc}
\toprule
\textbf{Mechanism} & \textbf{Architecture} & \textbf{Parameters} & \textbf{Pretrained Dataset} & \textbf{P} & \textbf{C} & \textbf{A} & \textbf{R} & \textbf{Avg} \\
\midrule
MoCov3 & ResNet-50 & 26.68M & ImageNet-1K  & 0.956886 &  0.808105 &  0.813993 & 0.782387
 & 0.840342 \\
SwAV & ResNet-50 & 26.68M & ImageNet-1K  & 0.958683 & 0.812012 & 0.799915 & 0.692034 & 0.815661 \\
DINO & ViT-S & 21.86M & ImageNet-1K  & 0.647904 & 0.557129 & 0.563567 & 0.516925 & 0.571381 \\
BEiT & ViT-B & 87.04M & ImageNet-21K  & 0.982635 & 0.905762 & 0.820392 & 0.812675 & 0.880366 \\
\bottomrule
\end{tabular}
\end{table*}

Table \ref{tab2} presents the aggregated performance of advanced deep learning models. Models pretrained on larger datasets, such as JFT-300M, ImageNet-21K, and ImageNet-22K, and evaluated on the Office-Home dataset exhibit varying accuracies across the : Product (P), Art (A), Clipart (C), and Real-world (R) domains. From the table, we can see that ResNetv2-101, pretrained on JFT-300M, achieved an average accuracy of 68.70\%. ViT models, such as ViT-S and ViT-B, pretrained on ImageNet-21K, scored average accuracies of 76.04\% and 73.78\%, respectively. ViT-S, being smaller (22.65M parameters), may be better at distilling generalizable features from the large-scale dataset without overfitting to irrelevant details, which allows it to outperform ViT-B when pretrained on ImageNet-21K. Meanwhile, Swin Transformers (Swin-S and Swin-B), pretrained on ImageNet-22K, demonstrated strong performance, with average accuracies of 81.35\% and 81.20\%, respectively. ConvNeXt models, particularly ConvNeXt-B, outperformed all others by achieving the highest average accuracy of 84.47\%, showcasing their effectiveness in domain generalization tasks on the Office-Home dataset. Despite having relatively fewer parameters, ConvNeXt-N performed well, achieving an average accuracy of 73.76\% when pretrained on the ImageNet-12K dataset. We were unable to obtain the performance of ResNet-18, ResNet-50, and ConvNeXt-A and ConvNeXt-P models when pretrained on larger datasets, as they are not publicly available.

Table \ref{tab3} presents a comprehensive comparison of several SSL methods using the ResNet-50, ViT-S, and ViT-B architectures. Models were pretrained on the ImageNet-1K and ImageNet-21K datasets and evaluated on the Office-Home dataset. The evaluated methods include MoCov3, SwAV, DINO, and BEiT. Both SwAV and MoCov3 SSL pre-trainings demonstrated well-balanced performance, achieving average accuracies of 65.50\% and 66.59\%, respectively. However, DINO exhibited significant underperformance, with a notably low average accuracy of 13.74\%. In contrast, the BEiT model achieved superior results, with an average accuracy of 78.12\%, outperforming across all domains. This highlights the robustness of BEiT in adapting to various domains through self-supervised learning.

Tables \ref{Tab1} and \ref{tab2} demonstrate that ConvNeXt models, excel in DG on the Office-Home dataset, achieving superior accuracy. As shown in Table \ref{tab3}, the BEiT pretraining performs well in SSL settings which indicates its strong ability to adapt across diverse domains. Fig. \ref{tnse01} shows the t-SNE plots for the BEiT pretraining, ConvNeXt-B pretrained on ImageNet-1K and  ImageNet-22K dataset as these models achieves highest performance individually from Tables \ref{Tab1},  \ref{tab2} and  \ref{tab3} respectively. From the figure, it is clear that ConvNeXt-B pretrained on ImageNet-22K exhibits better clustering and distinctiveness of features compared to other models, indicating a more effective feature space adaptation which potentially contributes to its superior performance in DG tasks.

\begin{figure*}[!t]
	\centerline{\includegraphics[width=16cm]{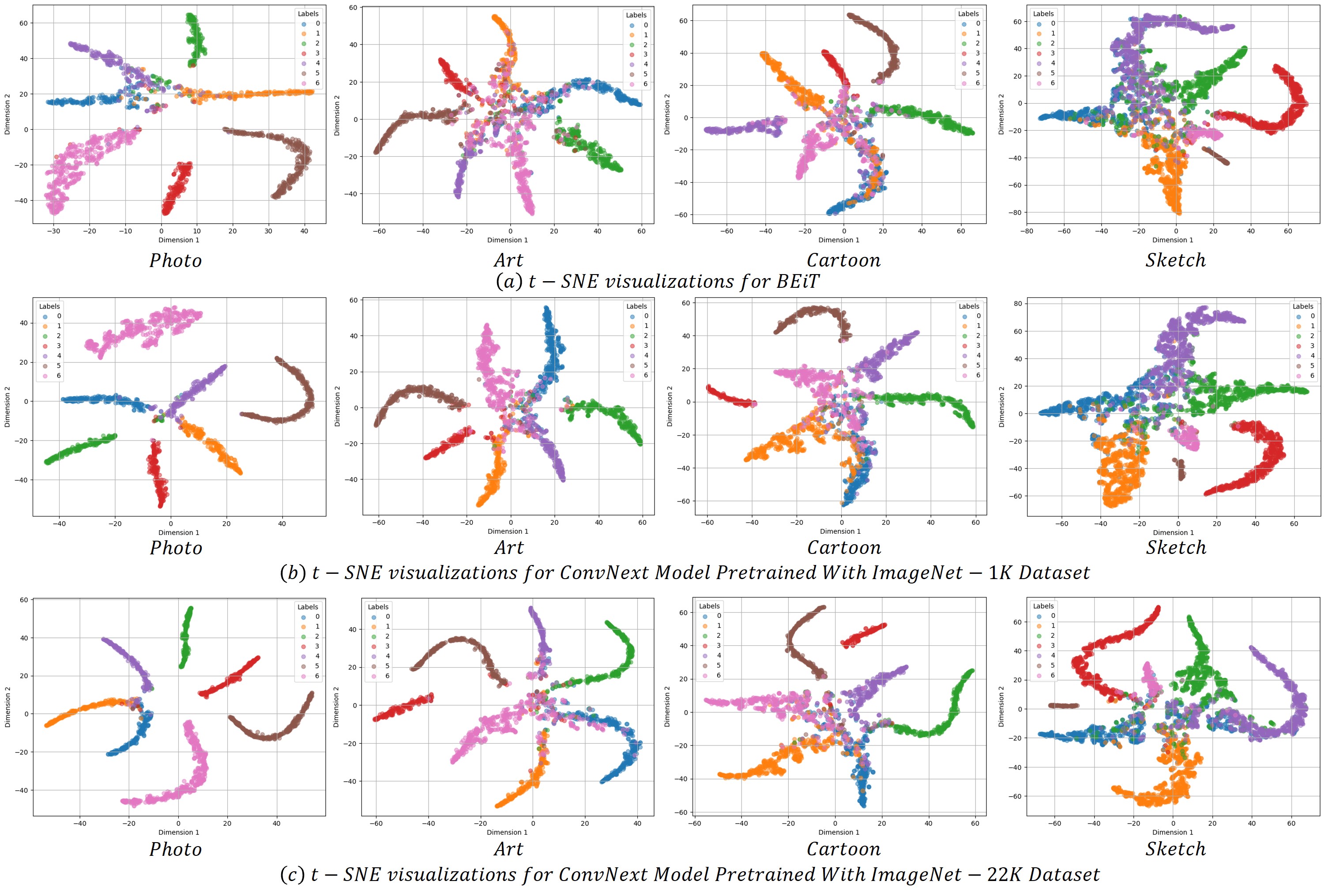}}
	\caption{t-SNE plots for feature representations from different models: BEiT pretraining, ConvNeXt-B pretrained on ImageNet-1K, and ConvNeXt-B pretrained on ImageNet-22K for the PACS dataset. The ConvNeXt-B model pretrained on ImageNet-22K demonstrates better clustering and distinctiveness of features compared to the other models, suggesting a more effective feature space adaptation, which likely contributes to its superior performance in DG tasks.}
	\label{tsne02}
\end{figure*}
\subsection{Results on PACS dataset}
The results on the PACS dataset follow similar trends as previously discussed, with the ConvNeXt architecture consistently outperforming other models. Table \ref{tab4} details the performance of models pre-trained on ImageNet-1k and evaluated across the four PACS domains: Photo (P), Art Painting (A), Cartoon (C), and Sketch (S). The ResNet models (ResNet-18, ResNet-50, and ResNetv2-101) show a progressive improvement in accuracy, with ResNetv2-101 achieving an average of 82.91\%. ViT models (ViT-S and ViT-B), along with the Swin Transformers (Swin-T, Swin-S, and Swin-B), demonstrate competitive results. ConvNeXt-B, however, outperforms all other models, achieving the highest overall accuracy of 90.73\%. The other variants of the ConvNeXt models exhibit relatively similar performance.

Table \ref{tab5} presents the performance metrics of various models pretrained on larger datasets and evaluated on the PACS dataset. The results show that ResNetv2-101, pretrained on JFT-300M, achieves an average accuracy of 83.99\%. Among the ViT models pretrained on ImageNet21k, ViT-S demonstrates higher accuracies, achieving average accuracies of 85.91\%. Swin Transformer models Swin-T, Swin-S and Swin-B, pretrained on ImageNet-22K, further enhance performance, demonstrating even greater robustness across all domains, with Swin-B reaching an average of 90.95\%. Notably, ConvNeXt models, especially ConvNeXt-B, pretrained on ImageNet-22K, surpass all other models, achieving the highest overall accuracy of 92.55\%, showcasing superior adaptability and performance across the diverse visual contexts of the PACS dataset.

Table \ref{tab6} illustrates a comparative analysis of four self-supervised methods, specifically MoCov3, SwAV, DINO, and BEiT, using the PACS dataset. The SwAV model attains an average accuracy of 81.57\%. DINO, on the other hand, demonstrates a notable lack of performance with an average accuracy of 57.14\%. BEiT surpasses both with the highest overall average accuracy of 88.04\%.

Fig. \ref{tsne02} shows the t-SNE plots for the BEiT pretraining, ConvNeXt-B pretrained on ImageNet-1K and  ImageNet-22K dataset as these models achieves highest performance individually from Tables \ref{tab4},  \ref{tab5} and  \ref{tab6} respectively. From the figure, it can be clearly observed that ConvNeXt-B pretrained on ImageNet-22K exhibits better clustering and distinctiveness of features compared to other models which indicates a more effective feature space adaptation which potentially contributes to its superior performance in FDG tasks.

\begin{figure}[!t]
	\centerline{\includegraphics[width=9cm]{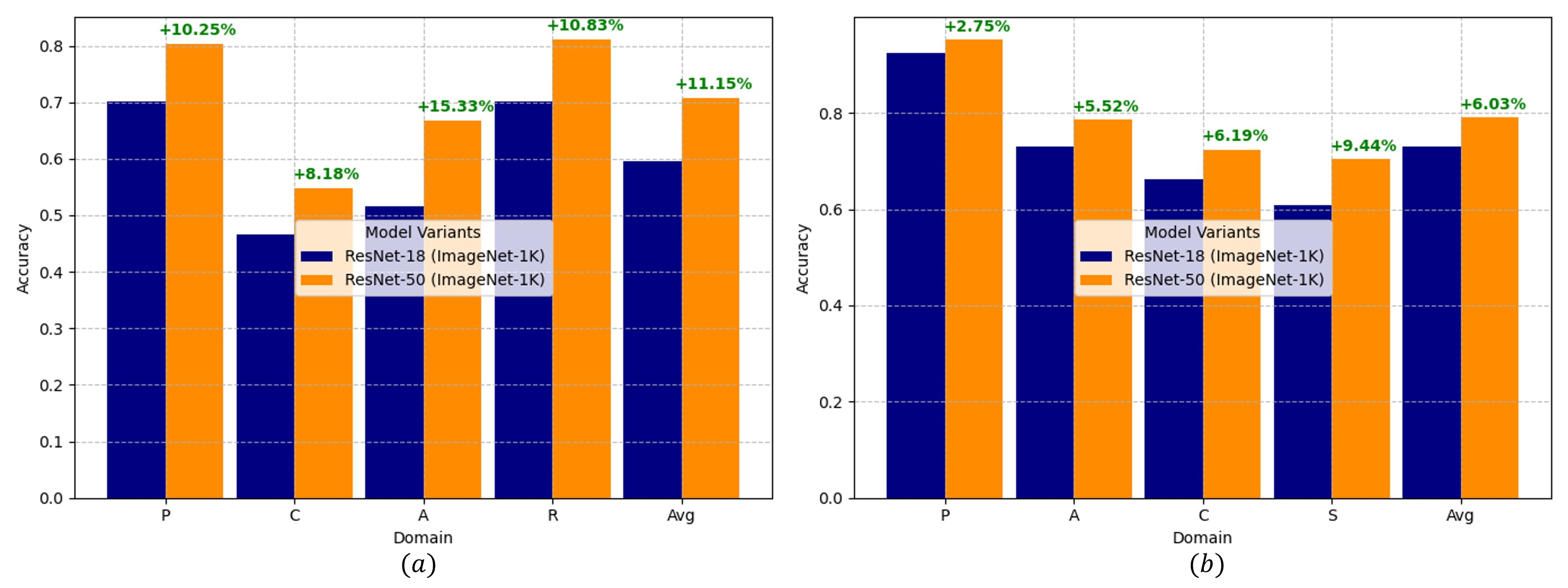}}
	\caption{Comparative Analysis of Architectural Effects of ResNet variants (ResNet-18 vs ResNet-50) on FDG Performance (a) Office-Home Dataset (b) PACS Dataset}
	\label{ad_res}
\end{figure}

\begin{figure}[!t]
	\centerline{\includegraphics[width=9cm]{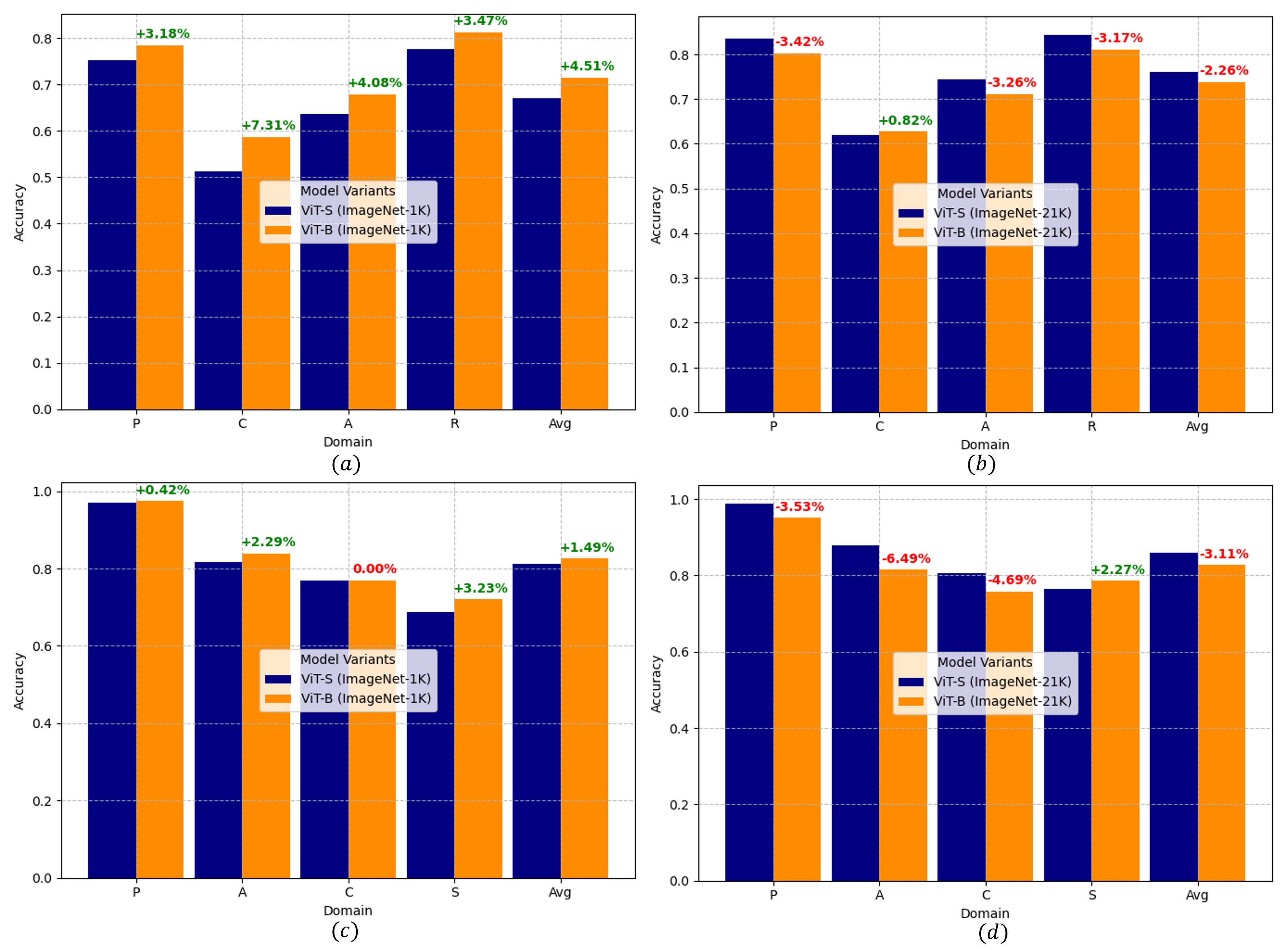}}
	\caption{Comparative Analysis of Architectural Effects on FDG Performance Between ViT Variants: This figure illustrates the performance differences of ViT-S and ViT-B architectures, analyzed across various configurations and datasets. Panels depict: (a) Office-Home dataset with models pre-trained on ImageNet-1K, (b) Office-Home dataset with models pre-trained on ImageNet-21K, (c) PACS dataset with models pre-trained on ImageNet-1K, (d) PACS dataset with models pre-trained on ImageNet-21K.}
	\label{ad_vit}
\end{figure}

\begin{figure}[!t]
	\centerline{\includegraphics[width=9cm]{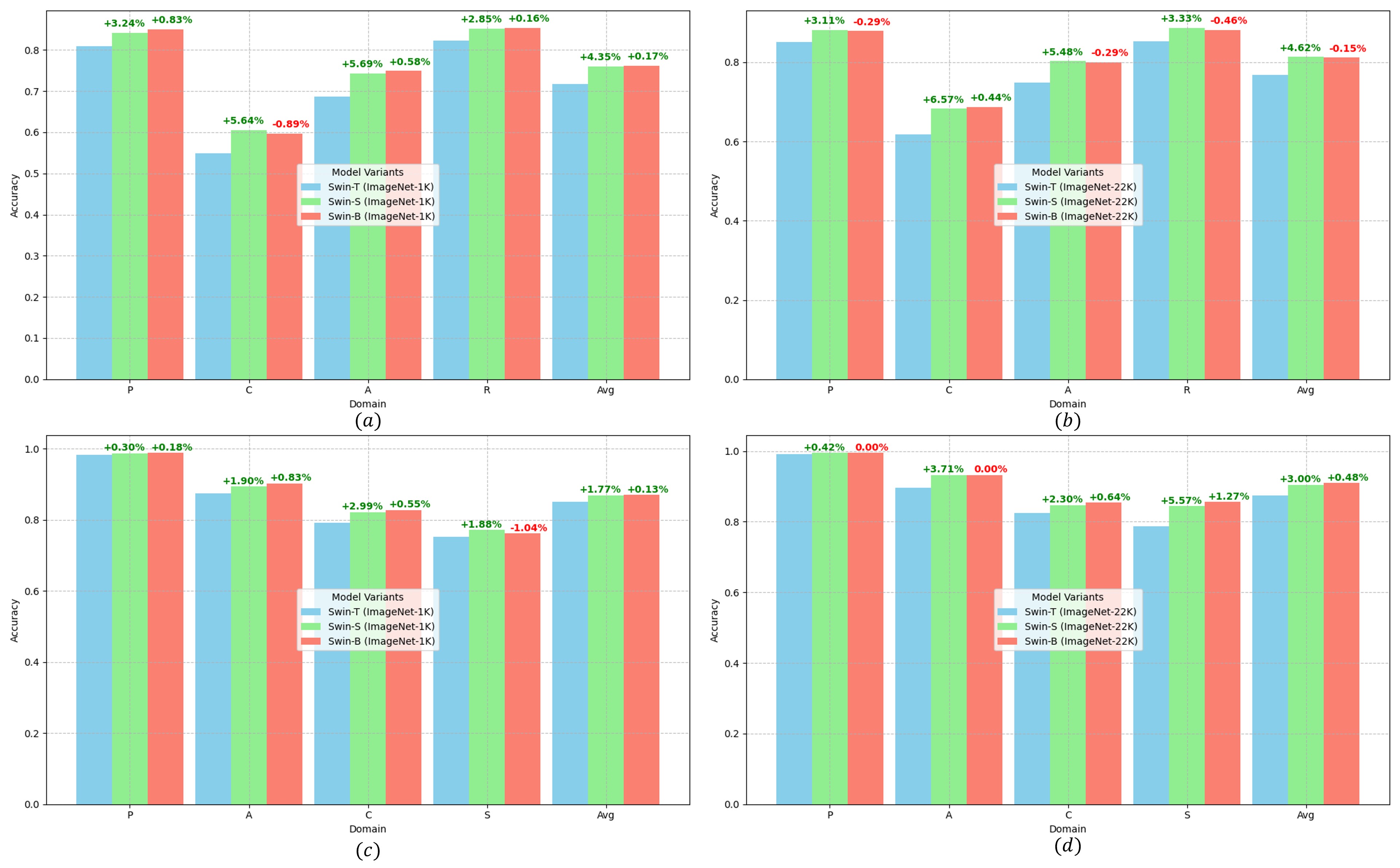}}
	\caption{Comparative Analysis of Architectural Effects on FDG Performance Among Swin Transformer Variants: This figure illustrates the performance differences of Swin-T, Swin-S, and Swin-B architectures, analyzed across various configurations and datasets. Panels depict: (a) Office-Home dataset with models pre-trained on ImageNet-1K, (b) Office-Home dataset with models pre-trained on ImageNet-22K, (c) PACS dataset with models pre-trained on ImageNet-1K, (d) PACS dataset with models pre-trained on ImageNet-22K.}
	\label{ad_swin}
\end{figure}

\begin{figure}[!t]
	\centerline{\includegraphics[width=9cm]{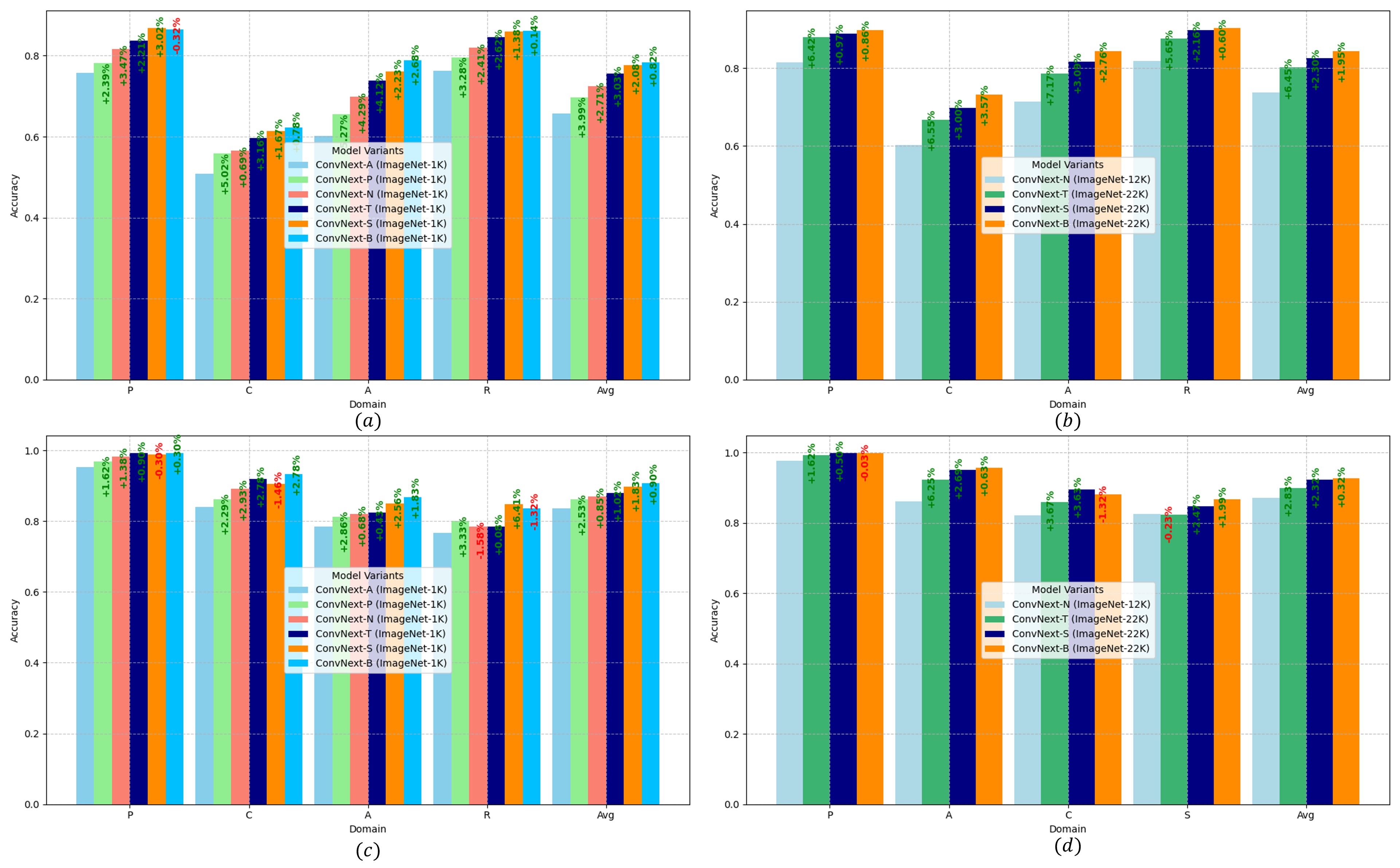}}
	\caption{Comparative Analysis of Architectural Effects on FDG Performance Among ConvNeXt Variants: This figure illustrates the performance differences of ConvNeXt-A, ConvNeXt-P, ConvNeXt-N, ConvNeXt-T, ConvNeXt-S and ConvNeXt-B architectures, analyzed across various configurations and datasets. Panels depict: (a) Office-Home dataset with models pre-trained on ImageNet-1K, (b) Office-Home dataset with models pre-trained on larger datasets, (c) PACS dataset with models pre-trained on ImageNet-1K, (d) PACS dataset with models pre-trained on larger datasets.}
	\label{ad_conv}
\end{figure}

\subsubsection{Self-Supervised vs Supervised Pretraining}
The results presented in Tables \ref{tab3} and \ref{tab6} highlight that the effectiveness of SSL methods—such as MoCov3, SwAV, DINO, and BEiT—is highly dependent on the nature of the downstream dataset. The key observation is that the performance of these SSL mechanisms varies significantly between datasets with subtle domain differences and those with more pronounced domain distinctions. On the Office-Home dataset, which features subtle domain variations (see Fig. \ref{tnse_domains}$(a)$), supervised pre-training generally outperforms SSL pre-trainings. For example, ResNet-50 with supervised pre-training achieves a mean accuracy of 70.75\%, whereas MoCov3 and SwAV attain lower accuracies of 66.60\% and 65.49\%, respectively. This suggests that SSL methods may struggle to generalize effectively when domain differences are minimal and nuanced, as they might not capture the slight variations between domains as well as supervised methods do.

In contrast, on the PACS dataset—characterized by more distinct and diverse domains (see Fig. \ref{tnse_domains}$(b)$)—SSL pre-trainings demonstrate superior performance. MoCov3 surpasses supervised pre-trainings by achieving an accuracy of 84.03\%, compared to 79.15\% for the supervised ResNet-50. SwAV also performs competitively with an accuracy of 81.56\%. These results indicate that SSL pre-trainings are particularly effective in environments with clear and significant domain differences, as they can leverage this diversity to learn more generalizable and robust features. However, not all SSL methods perform equally across datasets. DINO, despite utilizing a self-distillation mechanism, shows a notable performance gap on both PACS and Office-Home datasets. This underperformance may be due to over-specialization on specific augmentation patterns during pre-training, which limits its ability to generalize across different domains and reduces its effectiveness in domain generalization tasks.

On the other hand, the BEiT model, when pretrained on the ViT-B architecture, outperforms its supervised counterparts on both datasets. BEiT achieves an accuracy of 78.12\% on Office-Home and 88.04\% on PACS, compared to 73.78\% and 82.59\%, respectively, for supervised training. BEiT's approach involves reconstructing masked image patches, which encourages the model to focus on the intrinsic structure and semantics of images rather than domain-specific details. This enables BEiT to generalize better across varied domains by capturing more universal image representations.

These observations suggest that the choice between SSL and supervised pre-training should be informed by the characteristics of the target dataset. The Key takeaways can be summarized as:

\begin{itemize}
\item Datasets with Distinct Domains: In datasets like PACS, where domain diversity is significant, SSL pre-training methods like MoCov3 and SwAV excel by effectively learning from the domain differences. During pre-training, these methods learn representations that are invariant to various augmentations and transformations, capturing high-level semantic features that are less sensitive to domain-specific variations. This makes them particularly effective when fine-tuned on datasets with distinct domains, as they can leverage the diversity to learn more generalizable features, leading to better performance in domain generalization tasks.

\item Datasets with Subtle Domain Variations: In datasets such as Office-Home, where domains overlap and differences are subtle, supervised pre-training tends to perform better. Supervised pre-training leverages labeled data to learn discriminative features that can capture nuanced variations between classes and domains. This detailed feature representation is crucial when fine-tuning on datasets with subtle domain differences, as it allows the model to more precisely capture the slight variations between domains, which SSL methods might overlook due to their focus on learning invariant features.

\item Among the SSL techniques evaluated, BEiT stands out for its strong generalization capabilities across both distinct and overlapping domains. Its strategy of learning to reconstruct masked image patches during pre-training encourages the model to understand both local details and global context, leading to rich feature representations. This makes BEiT particularly robust when fine-tuned on a wide range of applications, as it captures intrinsic image structures rather than relying on domain-specific cues. Models leveraging BEiT's approach offer improved generalization across diverse domains, making them particularly suitable for tasks requiring robustness to domain variations.

\end{itemize}

\subsection{Architectural Depth Effects}
Figs.~\ref{ad_res}, \ref{ad_vit}, \ref{ad_swin}, and \ref{ad_conv} present a comparative analysis of the effects of architecture on FDG performance across the Office-Home and PACS datasets for ResNet, ViT, Swin Transformer, and ConvNeXt architectures, respectively.
It is evident that ResNet models are significantly more sensitive to architectural changes compared to advanced architectures. Specifically, ResNet-50 demonstrates superior performance over ResNet-18 by achieving average accuracy improvements of $11.15\%$ and $6.03\%$ on Office-Home and PACS dataset respectively. ViT models show performance variances between $4.51\%$ to $-3.11\%$ average accuracy when experimented with different architectures, reflecting their dependency on scale for capturing global dependencies effectively. From Figs.~\ref{ad_vit}$(a)$ and ~\ref{ad_vit}$(c)$, it is evident that the ViT-B model outperformed the ViT-S model when pre-trained on the ImageNet-1K dataset, achieving an accuracy improvement of 4.51\% on the Office-Home dataset and 1.49\% on the PACS dataset, respectively. However, when pre-trained on the ImageNet-21K dataset, the relatively smaller ViT-S model outperforms its larger counterpart, ViT-B. Specifically, ViT-S achieved accuracy improvements of 2.26\% on the Office-Home dataset and 3.11\% on the PACS dataset. This superior performance of ViT-S, which has 22.65 million parameters, can be attributed to its ability to more effectively distill generalizable features from the extensive ImageNet-21K dataset. The smaller model size helps prevent overfitting to less relevant details, enabling it to surpass the performance of ViT-B under these conditions.
\begin{figure}[!t]
	\centerline{\includegraphics[width=9cm]{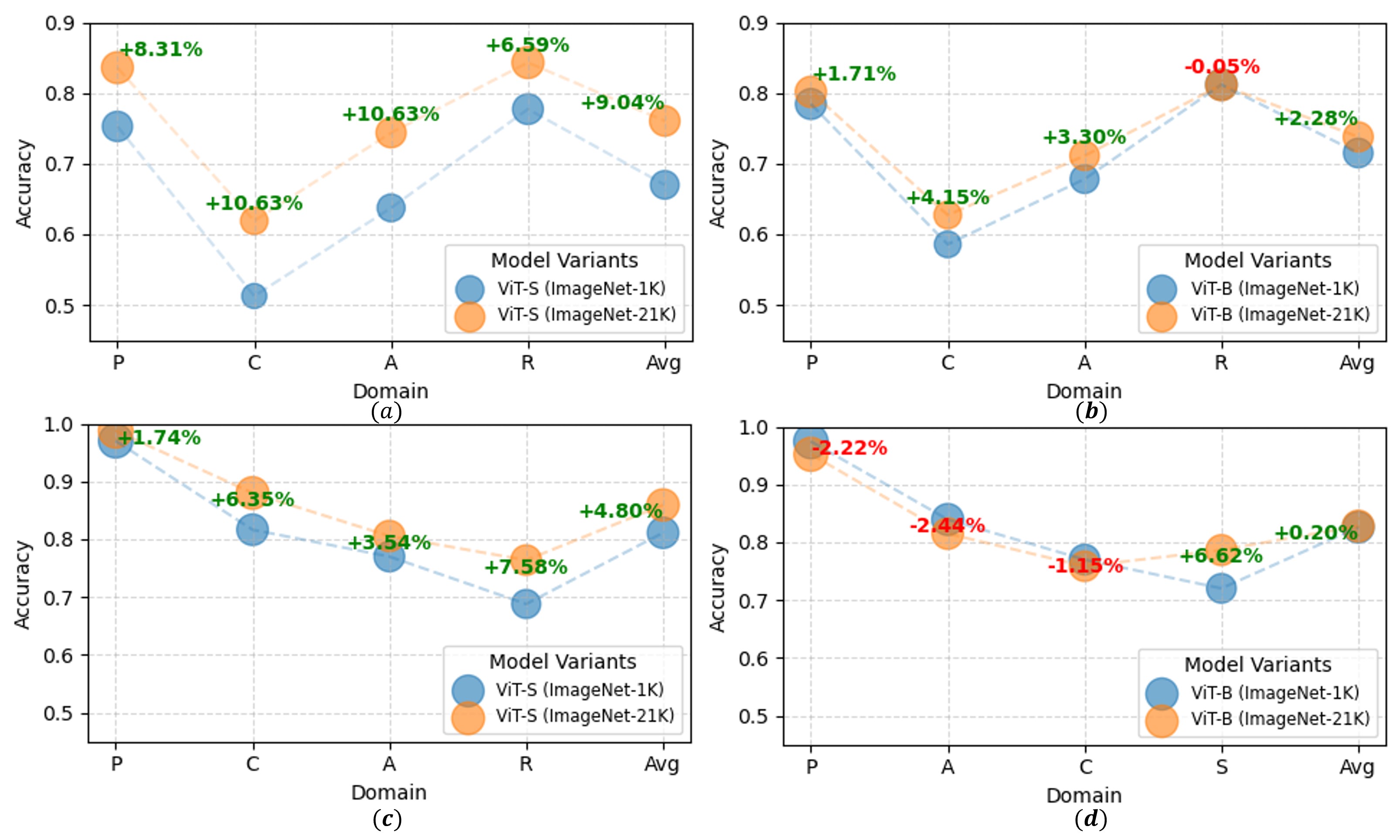}}
	\caption{Comparative Performance Analysis of the Effects of Pretraining Datasets (ImageNet-1K vs. ImageNet-21K) on ViT Models: (a) ViT-S Performance on the Office-Home Dataset; (b) ViT-B Performance on the Office-Home Dataset; (c) ViT-S Performance on the PACS Dataset; (d) ViT-B Performance on the PACS Dataset.}
	\label{DE_ViT}
\end{figure}
\begin{figure}[!t]
	\centerline{\includegraphics[width=9cm]{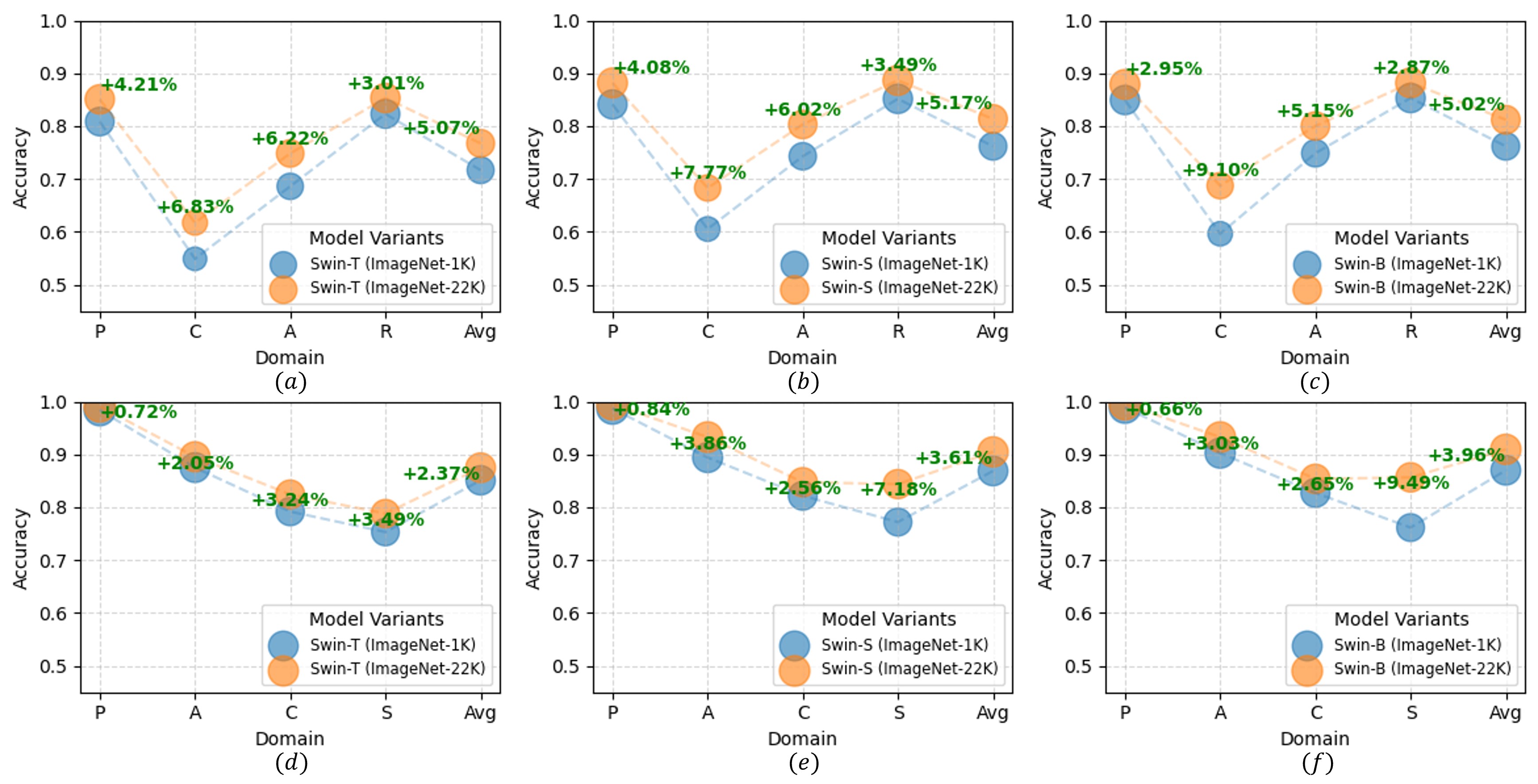}}
	\caption{Comparative Performance Analysis of the Effects of Pretraining Datasets (ImageNet-1K vs. ImageNet-22K) on Swin Transformers: (a) Swin-T Performance on the Office-Home Dataset; (b) Swin-S Performance on the Office-Home Dataset; (c) Swin-B Performance on the Office-Home Dataset; (d) Swin-T Performance on the PACS Dataset; (e) Swin-S Performance on the PACS Dataset; (f) Swin-B Performance on the PACS Dataset.}
	\label{DE_Swin}
\end{figure}
In contrast, Swin Transformer models show better robustness to architectural differences. From Fig.~\ref{ad_swin}, it is observable that for both the Office-Home and PACS datasets, the Swin-S and Swin-B variants exhibited comparable performance, irrespective of whether they were pre-trained on ImageNet-1K or ImageNet-22K. In contrast, the smaller Swin-T model experienced a drop in performance compared to the larger Swin-S and Swin-B. However, this performance drop is mitigated when Swin-T is pre-trained on the larger ImageNet-22K dataset. This suggests that the broader and more diverse training provided by ImageNet-22K helps enhance the generalization capabilities of smaller models like Swin-T, reducing the gap in performance with larger models.
\begin{figure*}[!t]
	\centerline{\includegraphics[width=16cm]{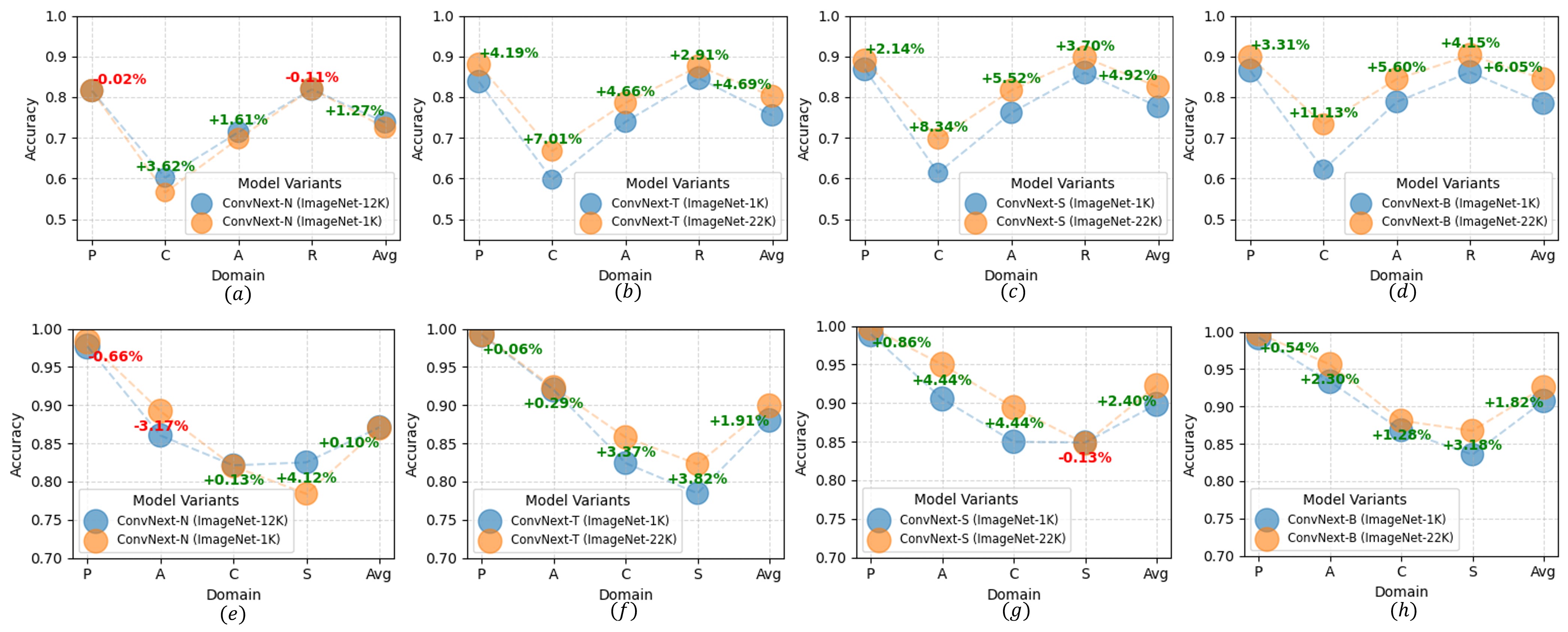}}
	\caption{Comparative Performance Analysis of the Effects of Pretraining Datasets on ConvNeXt Models: (a) ConvNeXt-N Performance on the Office-Home Dataset; (b) ConvNeXt-T Performance on the Office-Home Dataset; (c) ConvNeXt-S Performance on the Office-Home Dataset; (d) ConvNeXt-B Performance on the Office-Home Dataset; (e) ConvNeXt-N Performance on the PACS Dataset; (f) ConvNeXt-T Performance on the PACS Dataset; (g) ConvNeXt-S Performance on the PACS Dataset; (f) ConvNeXt-B Performance on the PACS Dataset.}
	\label{DE_Conv}
\end{figure*}
\begin{figure}[!t]
	\centerline{\includegraphics[width=9cm]{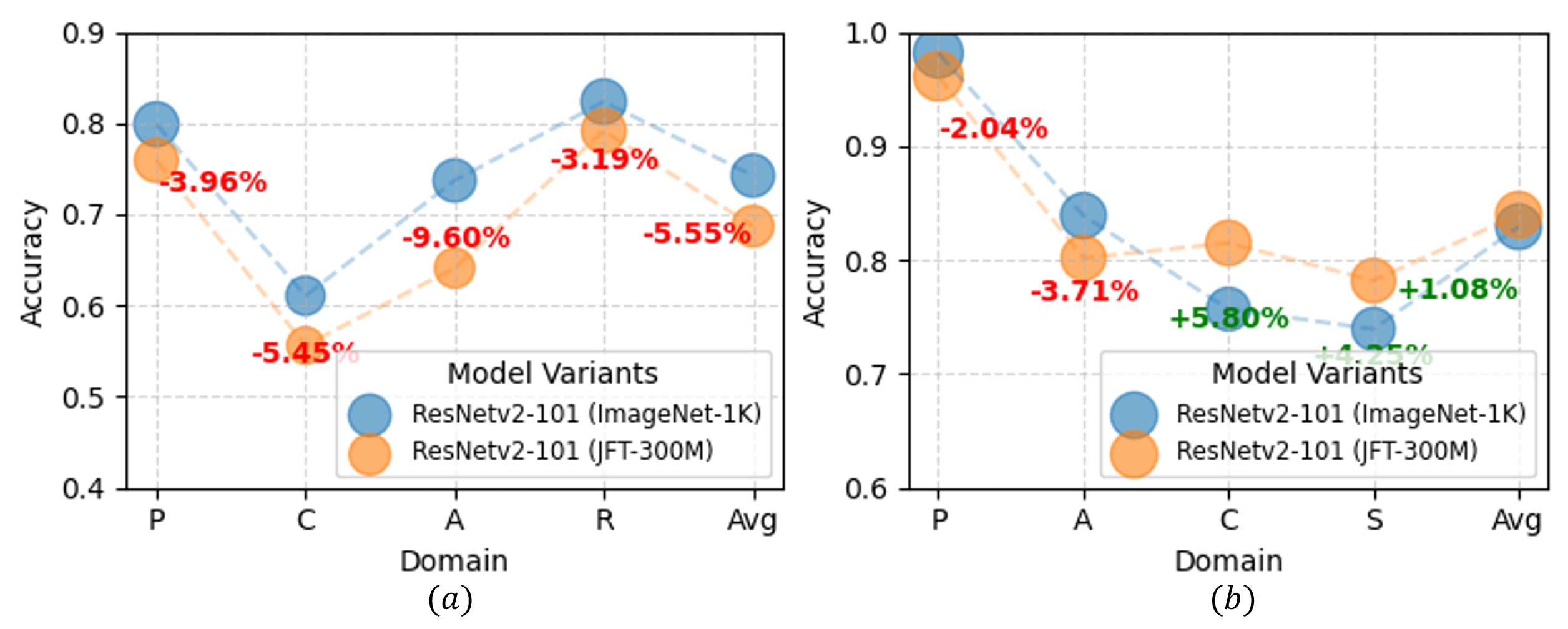}}
	\caption{Comparative Performance Analysis of the Effects of Pretraining Datasets (ImageNet-1K vs. JFT-300M) on ResNetv2-101 Model: (a) ResNetv2-101 Performance on the Office-Home Dataset; (b) ResNetv2-101 Performance on the PACS Dataset.}
	\label{DE_res}
\end{figure}
 \begin{figure*}[t]
	\centerline{\includegraphics[width=16cm]{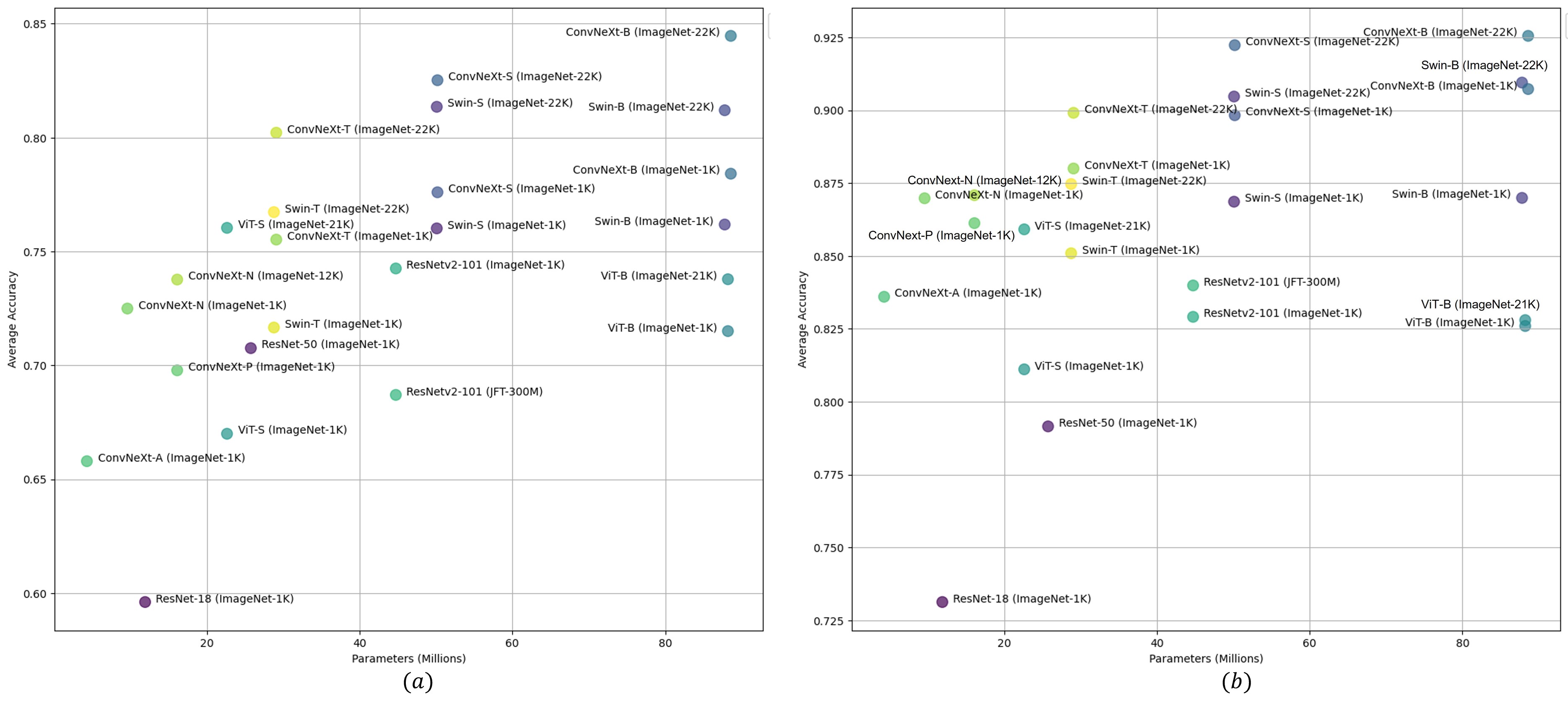}}
	\caption{Comparison of model parameters and accuracy across Office-Home and PACS datasets, highlighting the superior performance of models like Swin-S and ConvNeXt-S over ResNet variants.}
	\label{ParameterVsAcc}
\end{figure*}
Conversely, Fig.~\ref{ad_conv} reveals that the ConvNeXt models demonstrate a consistent enhancement in performance correlated with increases in model depth, indicating that larger architectures systematically improve average accuracy. Notably, the rate of accuracy improvement is more pronounced with the Office-Home dataset compared to the PACS dataset. This suggests that for complex datasets, where domain distinctions are less clear, deeper models contribute significantly to achieving higher accuracy.
\subsection{Pretraining Dataset Effects}
Figs.~\ref{DE_ViT}, \ref{DE_Swin}, \ref{DE_Conv} and \ref{DE_res} provide a detailed comparative analysis of the impact of pre-training datasets on FDG performance across the Office-Home and PACS datasets. For Figs.~\ref{DE_ViT}, \ref{DE_Swin}, \ref{DE_Conv}, the first row represents the performance in the Office-Home dataset, and the second row represents the performance in the PACS dataset. It is evident from these figures that the majority of the models, with one exception, exhibited enhanced accuracy when pre-trained on larger datasets. From Fig. ~\ref{DE_ViT}, we can observe that the smaller version of the ViT model achieved significant improvements irrespective of the downstream dataset. For the Office-Home and PACS datasets, the ViT-S achieved an average accuracy improvement of 9.04\%  and 4.80\%, respectively. However, the larger version, ViT-B, achieved an average improvements of 2.28\% and 0.20\% when pre-trained with a larger ImageNet-21K dataset. 

The Swin Transformer and ConvNeXt models, on the other hand, demonstrated very similar performance across their respective B, S, and T variants. Both models achieved significant improvements when pre-trained with larger datasets. However, for the Office-Home dataset, these models exhibited relatively greater performance gains compared to the PACS dataset. The ConvNeXt-N version, with its simpler architecture, achieved only modest performance improvements when pre-trained on the smaller ImageNet-12K dataset. This outcome may be partly due to the limited size of the pretraining dataset; with ImageNet-12K being smaller than ImageNet-22K, the performance improvements are not as substantial. Pretraining on a larger dataset like ImageNet-22K could potentially enhance the model's ability to exploit the full complexity and variety of the data, leading to greater performance gains.

The ResNetv2-101 model also showed an improvement, albeit more modest at 1.08\%, when pre-trained on the JFT-300M dataset. However, for the Office-Home dataset, the performance of ResNetv2-101 declined. Collectively, the findings indicate that models pre-trained on larger datasets generally yield superior performance. Notably, these models demonstrate more substantial performance gains on the Office-Home dataset compared to the PACS dataset. This discrepancy can be attributed to the greater complexity of the Office-Home dataset, which is characterized by more significant domain feature overlap. Models that undergo pre-training on extensive datasets encounter a wider variety of features and variations, equipping them with enhanced generalization capabilities. Consequently, when these models are fine-tuned on a multifaceted dataset like Office-Home, they achieve more pronounced accuracy improvements.

\subsection{Parameter Effects} 
Fig. \ref{ParameterVsAcc} effectively depicts the relationship between the number of parameters in various models and their mean accuracy, emphasizing the differences in model performance across the Office-Home and PACS datasets, as presented in Figs. \ref{ParameterVsAcc} $(a)$  and $(b)$, respectively. The models can be divided into four parameter ranges: tiny (0-20M), low (20-30M), medium (40-50M), and large (85-90M). A correlation can be observed where increased parameters generally lead to higher accuracy in both datasets. Models like Swin-S and ConvNeXt-S, which have undergone advanced pre-training on the ImageNet-22K dataset, undeniably showcase superior performance. 

When comparing these models to ResNet variants, it becomes clear why more advanced architectures like ConvNeXt, ViT, and Swin-T should be prioritized. These newer models, even with fewer parameters, consistently outperform ResNet-18 and ResNet-50. For instance, ConvNeXt-A and ConvNeXt-P, with only 4.31M and 9.61M parameters, respectively, achieve better accuracy than ResNet-18, despite the latter's 11.91M parameters. This highlights the efficiency of ConvNeXt's architecture in leveraging modern training techniques and optimization. Similarly, models like ViT-S and Swin-T, which have significantly fewer parameters than larger ResNet variants such as ResNet-50 and ResNetv2-101, also demonstrate superior performance. Advanced architectures like ConvNeXt, ViT, and Swin Transformers are designed to excel in complex feature learning, making them ideal for tasks that require high levels of generalizability. These models not only offer higher accuracy but also a more efficient use of computational resources, making them ideal candidates for deployment in both high-performance and resource-constrained environments.


\section{Conclusion} \label{conclusion}
In this study, we explored the application of advanced pre-trained architectures such as  ViT, ConvNeXt, and Swin Transformer within the framework of FDG. Our work highlights the significant potential of these next-generation models to enhance the adaptability and generalization of FL systems across diverse and unseen domains while maintaining data privacy. By leveraging extensive pre-training datasets such as ImageNet-1K, ImageNet-21K, JFT-300M, and ImageNet-22K, we demonstrated the advantages of utilizing large-scale and diverse data sources in improving model performance.
Our experimental results show that the ConvNeXt model, particularly when pre-trained on ImageNet-22K, achieves superior accuracy, setting a new benchmark for FDG with 84.46\% on the Office-Home dataset and 92.55\% on the PACS dataset. The Swin Transformer also exhibits commendable performance, further underscoring the effectiveness of employing advanced architectures in FL scenarios. These findings not only confirm the robustness of these architectures in handling domain shifts but also emphasize the importance of sophisticated pre-training strategies in advancing FL capabilities.
The comparative analysis of self-supervised and supervised pre-training approaches provides valuable insights into their distinct impacts on FDG, offering a foundation for future research to refine and optimize training methodologies. Overall, our research establishes a new standard for FDG, highlighting the critical role of cutting-edge architectures and diverse pre-training datasets in pushing the boundaries of FL, thus paving the way for more robust and generalizable models in privacy-preserving collaborative learning environments.

\bibliographystyle{IEEEtran}
\bibliography{references}



\end{document}